\pdfoutput=1

\documentclass{article}

\PassOptionsToPackage{numbers, compress}{natbib}

\usepackage[preprint]{neurips_2025}

\usepackage[utf8]{inputenc} 
\usepackage[T1]{fontenc}    
\usepackage[colorlinks,citecolor=blue]{hyperref}       
\usepackage{url}            
\usepackage{lipsum}         
\usepackage{enumitem}
\usepackage{amsfonts}       
\usepackage{amssymb}
\usepackage{mathtools}
\usepackage{amsthm}
\usepackage{bm,verbatim}
\usepackage{bbm}
\usepackage{physics}

\usepackage{nicefrac}       
\usepackage{microtype}      
\usepackage[dvipsnames]{xcolor}

\usepackage{multirow}
\usepackage{tcolorbox}
\usepackage{wrapfig}
\usepackage{booktabs}
\usepackage{colortbl}
\usepackage{amsmath}

\usepackage{xpatch}
\makeatletter
\xapptocmd{\NAT@bibsetnum}{\setlength{\leftmargin}{0pt}\setlength{\itemindent}{\labelwidth}\addtolength{\itemindent}{\labelsep}}{}{}
\makeatother

\begin{document}

\title{T2I-Eval-R1: Reinforcement Learning-Driven Reasoning for Interpretable Text-to-Image Evaluation}

%

\author{
  \textbf{Zi-Ao Ma\textsuperscript{1}\footnotemark[1]},
  \textbf{Tian Lan\textsuperscript{1}\footnotemark[1]},
  \textbf{Rong-Cheng Tu\textsuperscript{2}},
  \textbf{Shu-Hang Liu\textsuperscript{1}},
\\
  \textbf{Heyan Huang\textsuperscript{1}},
  \textbf{Zhijing Wu\textsuperscript{1}},
  \textbf{Chen Xu\textsuperscript{3}\footnotemark[2]},
  \textbf{Xian-Ling Mao\textsuperscript{1}\footnotemark[2]}
\\
  \textsuperscript{1}School of Computer Science and Technology, Beijing Institute of Technology, China
\\
  \textsuperscript{2}Nanyang Technological University, Singapore
\\
  \textsuperscript{3}School of Medical Technology, Beijing Institute of Technology, China
\\
    \texttt{\{maziaoylwt,lantiangmftby\}@gmail.com,} \texttt{rongcheng.tu@ntu.edu.sg}
\\
    \texttt{chenxu05037@gmail.com},
    \texttt{maoxl@bit.edu.cn}
\\
\url{https://github.com/maziao/T2I-Eval-R1}
}

\renewcommand{\thefootnote}{\fnsymbol{footnote}}

\footnotetext[1]{\quad Equal contributions}
\footnotetext[2]{\quad Co-corresponding authors}

\maketitle

\begin{abstract}
The rapid progress in diffusion-based text-to-image (T2I) generation has created an urgent need for interpretable automatic evaluation methods that can assess the quality of generated images, therefore reducing the human annotation burden.
To reduce the prohibitive cost of relying on commercial models for large-scale evaluation, and to improve the reasoning capabilities of open-source models, recent research has explored supervised fine-tuning (SFT) of multimodal large language models (MLLMs) as dedicated T2I evaluators. However, SFT approaches typically rely on high-quality critique datasets, which are either generated by proprietary LLMs—with potential issues of bias and inconsistency—or annotated by humans at high cost, limiting their scalability and generalization.
To address these limitations, we propose T2I-Eval-R1, a novel reinforcement learning framework that trains open-source MLLMs using only coarse-grained quality scores, thereby avoiding the need for annotating high-quality interpretable evaluation rationale. Our approach integrates Group Relative Policy Optimization (GRPO) into the instruction-tuning process, enabling models to generate both scalar scores and interpretable reasoning chains with only easy acessible annotated judgment scores or preferences. Furthermore, we introduce a continuous reward formulation that encourages score diversity and provides stable optimization signals, leading to more robust and discriminative evaluation behavior.
Experimental results on three established T2I meta-evaluation benchmarks demonstrate that T2I-Eval-R1 achieves significantly higher alignment with human assessments and offers more accurate interpretable score rationales compared to strong baseline methods.
\end{abstract}

\vspace{-0.2cm}
\section{Introduction}
\vspace{-0.2cm}

Recent advancements in diffusion models have significantly improved the quality of text-to-image (T2I) generation, enabling the synthesis of high-resolution, photorealistic images from natural language prompts. Foundational models such as SD1.5~\cite{rombach2022high} and SDXL~\cite{podellsdxl} have pushed the frontier of generative fidelity and efficiency, while recent transformer-based architectures~\cite{peebles2023scalable} further enhance visual realism. Despite these advances, T2I systems still suffer from semantic inconsistencies—such as object hallucinations, attribute entanglement, and misalignment with complex or compositional instructions~\cite{gupta2025simple, esser2024scaling}—posing challenges for both quality control and model development.

These limitations underscore the pressing need for robust, fine-grained evaluation methods that can assess the semantic faithfulness of generated images to input prompts. However, standard evaluation metrics remain inadequate. Inception Score~\cite{salimans2016improved} and FID~\cite{heusel2017gans} capture only global fidelity and distributional similarity, while CLIPScore~\cite{hessel2021clipscore} offers a coarse proxy for image-text alignment using pretrained encoders but lacks interpretability and correlates poorly with human judgments.

To overcome these shortcomings, recent studies have explored leveraging large language models (LLMs) and multimodal LLMs (MLLMs) \cite{lu2023llmscore,hu2023tifa,vqascore,ku2024viescore,tu2024automatic} as evaluators, owing to their strong semantic understanding and reasoning abilities~\cite{ku2024viescore, lu2023llmscore}. Commercial MLLMs like GPT-4o have shown remarkable capability in scoring T2I outputs with chain-of-thought explanations, as demonstrated in frameworks such as VIEScore~\cite{ku2024viescore}. Nevertheless, relying on closed-source APIs introduces high cost, limited accessibility, and scalability concerns for real-world deployment.
Therefore, recent work has turned to training open-source MLLMs through supervised fine-tuning (SFT), aiming to replicate the evaluation ability of commercial models while avoiding their cost and constraints. For instance, EATG~\cite{tu2024automatic} formulates the evaluation task into sub-skills such as object presence, attribute correctness, and spatial alignment, and uses fine-grained supervision to train MLLMs to act as interpretable evaluators. 

However, this paradigm still relies on high-quality supervision dataset, which typically comes from either human annotations or synthetic critiques generated by proprietary models, such as GPT-4o \cite{achiam2023gpt} and Gemini \cite{gemini}. Human annotations are costly and time-consuming, requiring fine-grained judgments about object presence, attributes, and semantic alignment—making them difficult to scale across diverse prompts and domains. On the other hand, synthetic critiques often suffer from inconsistency, bias, and limited generalization, especially when transferred across models or applied to complex compositional prompts\cite{tu2024automatic}. These limitations significantly hinder the scalability and reliability of supervised fine-tuning for training open-source MLLMs as robust T2I evaluators.

To address these limitations, we propose T2I-Eval-R1, a novel framework that applies Rule-based Reinforcement Fine-Tuning (RFT) to train open-source MLLMs using only coarse-grained judgment scores or preferences. Unlike SFT approaches that depend on high-quality interpretable evaluation rationales, our method optimizes a reward-driven policy that encourages interpretable reasoning and semantic discrimination. Specifically, we integrate Group Relative Policy Optimization (GRPO) into the instruction-tuning pipeline to induce relative ranking behavior and structured rationales under weak supervision. Furthermore, we introduce a continuous reward formulation that increases reward diversity within training groups, improving learning stability and evaluator sharpness.
We validate T2I-Eval-R1 on the T2I-Eval~\cite{tu2024automatic}, TIFA v1.0~\cite{hu2023tifa}, and the ImageReward human preference dataset~\cite{xu2023imagereward} and demonstrate new state-of-the-art Spearman and Kendall correlations with human judgments. Our method not only surpasses strong open-source and GPT-4o-based baselines, but also provides interpretable rationales for each prediction, facilitating transparency and human trust.

\vspace{-0.2cm}
\section{Related Work}
\vspace{-0.1cm}
\subsection{Evaluation Methods for Synthesized Images}
\vspace{-0.1cm}

Early evaluation of text-to-image (T2I) generation focused on global, distributional metrics that trade interpretability for efficiency. Inception Score (IS) quantifies image quality and diversity via class‐prediction confidence and variety from a pretrained Inception‐v3 network~\cite{salimans2016improved}, while Fréchet Inception Distance (FID) measures the difference in mean and covariance between real and generated image embeddings~\cite{heusel2017gans}. Although these metrics are computationally cheap and widely used, they do not assess semantic fidelity to the input text and provide no insight into specific errors.

To incorporate semantic alignment, joint vision–language embeddings were introduced. CLIPScore computes the cosine similarity between CLIP‐encoded image and text representations to approximate text–image alignment without references~\cite{hessel2021clipscore}, and BLIPv2Score refines this approach with richer pretrained multimodal representations~\cite{li2023blip}. Despite improved alignment, these methods remain opaque, yielding a single score without explanatory reasoning.

Recognizing the need for interpretability, subsequent work leveraged large language models (LLMs) and visual question answering (VQA). LLMScore decomposes prompts into subquestions answered through text‐only reasoning, enabling compositional fidelity checks~\cite{lu2023llmscore}, while TIFA frames evaluation as VQA over prompt‐derived questions to deliver fine‐grained faithfulness via QA accuracy~\cite{hu2023tifa}. These task‐decomposed strategies provide more transparency but incur high computational overhead and rely on handcrafted pipelines.

More recently, multimodal LLMs (MLLMs) have been applied directly as evaluators, generating natural‐language rationales alongside scalar judgments. For example, VIEScore prompts GPT-4o to produce chain‐of‐thought critiques that align closely with human preferences~\cite{ku2024viescore}. Although effective, these proprietary‐API–based systems face challenges in cost, scalability, and reproducibility.

Open-source alternatives seek to overcome these barriers via supervised fine-tuning (SFT). T2I-Eval trains MLLMs on human‐annotated or LLM‐generated critiques, decomposing evaluation into sub‐tasks such as object presence, attribute correctness, and spatial alignment~\cite{tu2024automatic}. While this yields interpretable outputs, the dependence on large‐scale, fine-grained annotations limits scalability and may introduce synthetic biases.

Overall, evaluation has progressed from coarse statistical metrics to semantically rich, interpretable methods, yet existing approaches struggle to reconcile annotation cost, model accessibility, and explanation quality. This motivates methods that leverage weak or coarse‐grained supervision to achieve both interpretability and scalability.

\subsection{Reinforcement Learning for Evaluation}

The prohibitive cost and limited scalability of supervised fine-tuning on fine-grained critiques have driven a shift toward reinforcement learning (RL) methods that learn from weaker, more abundant signals.
At the core of many recent RL evaluators lies Direct Preference Optimization (DPO)~\cite{rafailov2023direct}, which treats a language model itself as the reward model and optimizes it directly on human pairwise rankings. The original DPO formulation showed that this approach can recover strong reward models without separately training a critic network. Building on this idea, UnifiedReward applies DPO to multimodal data—training on large‐scale human comparisons over images (and even video) to produce either pairwise preference judgments or pointwise scores with minimal annotation overhead~\cite{wang2025unified}. However, its outputs remain opaque scalar utilities, providing no insight into why one image is ranked above another.

An alternative RL paradigm, Group Relative Policy Optimization (GRPO), was introduced in the DeepSeek-R1 work to elicit reasoning behaviors in open LLMs by sampling groups of candidate outputs and updating the policy based on their relative rewards—thereby removing the need for a separate value network and enhancing stability~\cite{guo2025deepseek}. Subsequent work in multimodal reasoning has extended GRPO with richer advantage estimators and auxiliary critics, demonstrating its effectiveness for various tasks~\cite{huang2025vision,zhang2025r1}. For example, UnifiedReward-Think distills chain-of-thought rationales from GPT-4o on a seed preference set and then refines its multimodal policy via rejection sampling and GRPO~\cite{wang2025unifiedmultimodal}. The resulting model emits free-form explanations alongside each preference decision, improving transparency—but it still relies exclusively on pairwise data, limiting its ability to make use of single-wise evaluation datasets or perform single-image scoring. Generally speaking, to date these GRPO-based efforts have focused on discrete correctness or abstract reasoning benchmarks rather than on semantic text-to-image evaluation, and they still employ binary or thresholded reward schemes that fail to differentiate near-misses on a graded scale.

\section{Method~\label{sec:method}}

\subsection{Task Formulation\label{subsec:task_formulation}}

We denote by $P$ a natural-language text prompt and by $I$ (or $(I_{A}, I_{B})$) one or two images generated by a text-to-image generative model in response to $P$.  Let $D$ be the set of evaluation dimensions, and let $G$ be the corresponding set of guidelines that define how each dimension should be interpreted (e.g., numeric ranges, anchor descriptions, or relational criteria).  We define an evaluator:
\begin{equation}
    E:\; (P,\; I\;\text{or}\;(I_A,I_B),\; D,\; G)\;\longmapsto\;(r,\; q)
\end{equation}
where the outputs are:
(1) $r$, a chain-of-thought rationale in natural language;
(2) $q$, a quantitative judgment, which may take one of two forms: 
a single-image score $s \in [s_{\min}, s_{\max}]$ for single-wise evaluation;
or a preference choice $c \in \{A, B, T\}$ ($T$ for `tie') for pairwise comparison.

\subsubsection{Single-Wise Evaluation}
For perceptual-quality dimensions $D_{PQ}$ (e.g.,sharpness, color fidelity)—which concern properties intrinsic to the image $I$ and do not typically depend on the prompt—the evaluator produces:
\begin{equation}
    (r,\; s)\;=\;E\bigl(I,\;D_{PQ},\;G\bigr),
\end{equation}
where $s\in[s_{\min},s_{\max}]$.  Likewise, for semantic-consistency dimensions $D_{SC}$ (e.g.,object presence, attribute correctness, relational alignment)—which measure alignment between $P$ and $I$—the evaluator can be written as:
\begin{equation}
    (r,\; s)\;=\;E\bigl(P,\;I,\;D_{SC},\;G\bigr).
\end{equation}

By distinguishing these two orthogonal axes, our formulation subsumes both standalone image quality metrics and prompt-conditioned alignment checks within a single, unified operator $E$.

\subsubsection{Pairwise Evaluation}
When comparing two images $I_A$ and $I_B$ generated from the same prompt $P$, traditional preference models often yield a discrete choice $c\in\{A,B,T\}$.  We extend this to a continuous preference confidence by defining:
\begin{equation}
    (r,\; p_{\mathrm{conf}})\;=\;E\bigl(P,\;I_A,\;I_B,\;D,\;G\bigr),  
\quad p_{\mathrm{conf}}\in[0,1],
\end{equation}
where $p_{\mathrm{conf}} = 1.0$ indicates definitive preference for $I_A$, $0.0$ for $I_B$, $0.5$ denotes comparable quality, and intermediate values reflect uncertain differences.  This continuous encoding captures nuanced degrees of comparative quality and supports downstream differentiation during model optimization.

\subsubsection{Modularity and Generality}
Since single-vs-pairwise and perceptual-vs-semantic represent two independent binary choices, our framework defines a $2\times 2$ taxonomy of evaluation modes.
By parameterizing $E$ on $D$ and $G$, we enable:
\begin{itemize}
    \item \textbf{Arbitrary dimension sets}: New criteria (e.g.,aesthetic style, prompt-specific object counts) can be introduced simply by augmenting $D$ and supplying appropriate guidelines in $G$, without modifying $E$’s core logic.
    \item \textbf{Flexible input configurations}: Both single-wise and pairwise evaluations are handled by the same functional form.
    \item \textbf{Interpretable outputs}: The chain-of-thought $r$ provides transparent justification for each numeric judgment $q$.
\end{itemize}

In practice, we implement $E$ via a templated prompt construction that instantiates $P$, $I$ or $(I_A,I_B)$, the chosen $D$, and the instructions $G$ into a single query for a MLLM.  The model then returns $(r,q)$, yielding both an explanation and its corresponding scalar or confidence score. This formalism lays the foundation for our subsequent optimizing objective, dataset construction, and prompt-design procedures.

\subsection{Optimizing Objective~\label{subsec:objective}}

\paragraph{Pairwise Ranking Objective}
Early LLM-based reward models are trained on pairwise human preference data using a ranking objective derived from the Bradley–Terry formulation~\cite{bradley1952rank}.  Concretely, given an input $x$ (which may include a prompt and one or two images) and two candidate responses $y_c$ (chosen) and $y_r$ (rejected), the model $r$ with parameter $\theta$ is optimized to satisfy
\begin{equation}
    \mathcal{L}_{\mathrm{ranking}}
= -\log \sigma\bigl(r_\theta(x,y_c) - r_\theta(x,y_r) - m(r)\bigr)
\;+\;\bigl(r_\theta(x,y_c) + r_\theta(x,y_r)\bigr)^2,
\end{equation}
where $\sigma$ is the sigmoid function, $m(r)$ a preference margin~\cite{touvron2023llama}, and the $(r_\theta(x,y_c)+r_\theta(x,y_r))^2$ term centers the reward distribution~\cite{eisenstein2023helping}.  While this loss yields models capable of pairwise ranking, it does not endow them with the ability to generate interpretable rationales, as they output only scalar utilities without a chain-of-thought explanation. Apart from that, models using this objective heavily rely on pairwise preference data and are not able to make full use of single-wise annotated data during training, bringing about limitations to its broader applications to some extent.

\paragraph{MLE Objective}
By contrast, generative evaluators frame T2I evaluation as sequence prediction and are typically trained with a maximum likelihood estimation (MLE) loss. It can be adapted to both single-wise and pairwise evaluation by altering the evaluation prompt. Given an evaluation input $x$ and target token sequence $y=y_{1:T}$, the MLE objective is
\begin{equation}
    \mathcal{L}_{\mathrm{MLE}}
= -\sum_{i=1}^T \log p_\theta\bigl(y_i \mid x,\,y_{<i}\bigr).
\end{equation}

This approach forces the model to reproduce exactly the reference rationale and score, heavily penalizing even minor deviations—e.g.\ scoring 8 when the reference is 7 is treated as equally wrong as scoring 3—thus undermining smooth credit assignment for near-correct outputs~\cite{welleck2020neural}.

\paragraph{T2I-Eval-R1 Objective}
To unify interpretability and robustness, we adopt Group Relative Policy Optimization (GRPO) as our primary training paradigm. GRPO extends PPO‐style policy optimization to outcome‐based, online updates without requiring a separate value network. Let $\pi_\theta$ be the current generative evaluator policy and $G$ the group size (i.e., number of sampled outputs per prompt $q$). We optimize
\begin{align}
    \mathcal{L}_\mathrm{GRPO}(\theta) & = - \mathbb{E} [q \sim P(Q), \{o_i\}_{i=1}^{G} \sim \pi_{\theta_\mathrm{old}}(O | q)] \notag \\
    & \frac{1}{G} \sum_{i=1}^G  \left[
        \min \left(
            \frac{\pi_\theta(o_{i} | q)}{\pi_{\theta_\mathrm{old}}(o_{i} | q )} A_{i},
            \mathrm{clip}\left( \frac{\pi_\theta(o_{i} | q)}{\pi_{\theta_\mathrm{old}}(o_{i} | q )}, 1 - \epsilon, 1 + \epsilon \right) A_{i}
        \right)
        -
        \beta \mathbb{D}_\mathrm{KL}(\pi_\theta \| \pi_\mathrm{ref})
    \right]
\end{align}
where $A_i$ are advantages normalized from bespoke rewards $\{r_i\}_{i=1}^G$, $\beta$ is a KL‐penalty coefficient, and $\pi_{\mathrm{ref}}$ is the fixed reference policy (often taken as the initial policy $\pi_{\theta_0}$)~\cite{guo2025deepseek}. This objective encourages stable updates while preserving interpretability via generative chain-of-thought outputs.

Most prior uses of GRPO in LLM reasoning tasks employ \textbf{binary (0/1) rewards}: an output is scored 1 if ``correct'' (e.g., solves a math problem) or 0 otherwise, treating all errors equally~\cite{guo2025deepseek,jin2025search}. Such a design works well when correctness is unambiguous—incorrect reasoning always yields a wrong final answer, or any non‐answer action (e.g., issuing a search query during reasoning) is hardly detrimental.  

However, text-to-image evaluation is inherently subjective and graded on a continuous scale.  Consider two predicted scores for a reference quality of 7 (on [0,10]): predicting 8 is closer and arguably preferable to the reference than predicting 3, yet a binary reward would penalize both equally.  To address this, we introduce \textbf{continuous rewards} that smoothly reflect distance from the reference:

\begin{align}
  \mathcal{R}_{\mathrm{single}}(s_{\mathrm{pred}}, s_{\min}, s_{\max}, s_{\mathrm{ref}})
  &= 1 \;-\; 2\,\frac{\bigl|\mathrm{clip}(s_{\mathrm{pred}},s_{\min},s_{\max}) - s_{\mathrm{ref}}\bigr|}{s_{\max}-s_{\min}}, \\[6pt]
  \mathcal{R}_{\mathrm{pair}}(p_{\mathrm{pred}}, p_{\mathrm{ref}})
  &= 1 \;-\; 2\,\bigl|\mathrm{clip}(p_{\mathrm{pred}},0,1) - p_{\mathrm{ref}}\bigr|,
\end{align}
which map deviations into $[-1,1]$, providing finer gradients when predictions are near the reference.
We call this innovative GRPO objective with continuous reward \textbf{T2I-Eval-R1} objective.

By replacing brittle MLE and opaque ranking losses with T2I-Eval-R1 objective, MLLM-based evaluator learns to produce both accurate scalar judgments and coherent rationales across single-wise and pairwise settings without expensive fine-grained annotation.  

\subsection{Training Datasets\label{subsec:training_dataset}}

To evaluate our T2I-Eval-R1 objective under both single-wise and pairwise paradigms, we assemble two distinct training corpora by re-sampling and organizing existing public datasets.

\paragraph{Single-wise Evaluation Corpus}
From the original 13,000 text–image pairs in the T2I-Eval training set~\cite{tu2024automatic}, we randomly sample 9,000 examples for each of the three evaluation dimensions: appearance quality, intrinsic attribute consistency, relationship attribute consistency, and 9,000 examples for overall evaluation. By pairing each prompt–image instance with its corresponding dimension set and guidelines, we obtain a 36,000-sample dataset, and each sample is associated with a reference score ranging from 0 to 10. This balanced sampling ensures uniform coverage across dimensions during policy learning and avoids over-representation of any single aspect.

\paragraph{Pairwise Evaluation Corpus}
Our pairwise dataset is derived from the human preference rankings of ImageRewardDB~\cite{xu2023imagereward}, which assign each generated image a quality level from 1 (best) to 5 (worst). We construct image pairs whose rating‐level difference $\Delta r\in\{1,2,3,4\}$. Recognizing that pairs with small $\Delta r$ are inherently closer in quality while those with large $\Delta r$ are easier to distinguish, we weight samples in the ratio 1 : 2 : 2 : 1 for $\Delta r=1,2,3,4$, respectively. Within each $\Delta r$ subset, we balance positive pairs ($I_A$ rated better than $I_B$) and negative pairs in a 1 : 1 ratio. After resampling, this yields a training set with around 35,000 pairs.

Together, these two corpora provide robust, dimension-balanced supervision for both scalar scoring and continuous-confidence learning under our reinforcement-learning paradigm.
\subsection{Prompt Design~\label{subsec:prompt_design}}

To ensure both flexibility and interpretability, we assemble every evaluation prompt via a four-block template that instantiates the evaluator function $E$ (from Section~\ref{subsec:task_formulation}).  Each block is parameterized by the current evaluation mode (single vs. pair), the target dimensions $D$, and the associated guidelines $G$.  This design allows us to swap in new criteria or switch protocols without retraining the evaluator. The four blocks include:
(1) \textbf{Task Description}: Briefly states the evaluator's role and the active dimensions;
(2) \textbf{Annotation Inputs}: Provides the text prompt and one or two generated images;
(3) \textbf{Evaluation Guidelines}: Lists the dimension definitions and any scoring anchors;
(4) \textbf{Output Format}: Specifies exactly what the model must return, enclosed in distinct tags for parseability.

Because each segment is defined as a separate block, our template supports:
\begin{itemize}[nosep, leftmargin=25pt]
    \item \textbf{Dimension Swapping}: Swap in new $d_i$ and $G_i$ without retraining.
    \item \textbf{Protocol Switching}: Toggle between single-wise and pairwise modes without altering the template logic.
    \item \textbf{Range Adjustment}: Edit $[s_{\min},s_{\max}]$ based on actual requirements (e.g., 0–5, 1–100).
\end{itemize}

This modularity ensures that the evaluator can be rapidly re-targeted to novel evaluation tasks, maintaining both scalability and interpretability across diverse T2I evaluation scenarios. Please refer to Appendix~\ref{app:prompt_template} for more details about the prompt template.

\section{Experiments~\label{sec:experiment}}

\subsection{Experiment Setup~\label{subsec:exp_setup}}

\paragraph{Benchmark Datasets}
We assess our models trained with T2I-Eval-R1 objective on three public benchmarks, each targeting a distinct evaluation mode:
(1) T2I-Eval~\cite{tu2024automatic} targeting on single-wise evaluation and the same dimensions as the training set; 
(2) TIFA v1.0~\cite{hu2023tifa} targeting on single-wise evaluation yet an unseen faithfulness dimension;
(3) ImageReward~\cite{xu2023imagereward} targeting on pairwise evaluation.
Please refer to Appendix~\ref{app:benchmark} for more details.

\paragraph{Baseline Methods}
To contextualize our results, we compare our method against a variety of existing methods including FID~\cite{heusel2017gans}, CLIPScore~\cite{hessel2021clipscore}, ImageReward~\cite{xu2023imagereward} and VIEScore~\cite{ku2024viescore}, etc. Please refer to Appendix~\ref{app:baseline_methods} for more details.

\paragraph{Models}
Given the various datasets and evaluation protocols, we design several settings to train multiple models that can achieve optimal performance in different scenarios. The settings are:
\begin{itemize}[nosep, leftmargin=25pt]
    \item \textbf{T2I-Eval-R1}$_\textbf{Base}$: This model is trained with the full single-wise T2I-Eval dataset mentioned in Section~\ref{subsec:training_dataset}. This model can used for either single-wise evaluation or pairwise evaluation in the single-wise manner;
    \item \textbf{T2I-Eval-R1}$_\textbf{Enhance}$: On the basis of \textbf{T2I-Eval-R1}$_\textbf{Base}$, we further perform rejection sampling on the T2I-Eval dataset, resulting in a small enhanced dataset with 4,000 samples for further training. This model is an enhanced version to solve the harder evaluation tasks;
    \item \textbf{T2I-Eval-R1}$_\textbf{Pair}$: This model is trained with the full pairwise ImageRewardDB dataset mentioned in Section~\ref{subsec:training_dataset}. This model can be used only for pairwise evaluation;
    \item \textbf{T2I-Eval-R1}$_\textbf{General}$: To explore whether a model can be used for text-to-image evaluation in the single-wise and pairwise manners simultaneously, we trained the model with the mixed dataset of T2I-Eval and ImageRewardDB. This model can be adapted to any evaluation protocols defined in Section~\ref{subsec:task_formulation}.
\end{itemize}
Please refer to Appendix~\ref{app:training_settings} for more training details.

\subsection{Experimental Results}

According to the features of different variants of T2I-Eval-R1 series model, we conduct various evaluations for each of them and validate whether each of the settings can push the capability of the evaluators to the next level.
We first evaluated \textbf{T2I-Eval-R1}$_\textbf{Enhance}$ on the single-wise T2I-Eval benchmark, with the aim of verifying the ability of T2I-Eval-R1 objective to fit human preferences on given dimensions and comparing it with strong baseline methods. The experimental results are illustrated in Table~\ref{tab:overall_results}. According to these results, we can conclude the following insights:

\begin{table*}[htbp]
    \small
    \centering
    \vspace{-10pt}
    \caption{\label{tab:overall_results} Comparison of previous methods and ours on T2I-Eval benchmark, with top scores of each category in \textbf{bold} and top scores of all methods in \textcolor{red}{\textbf{red}}.}
    \resizebox{1.0\linewidth}{!}{
        \begin{tabular*}{0.98\linewidth}{lcccccccc}
        \toprule
             \multirow{2}{*}[-0.0pt]{\textbf{Method}} & \multicolumn{2}{c}{\textbf{Appearance}} & \multicolumn{2}{c}{\textbf{Intrinsic}} & \multicolumn{2}{c}{\textbf{Relationship}} & \multicolumn{2}{c}{\textbf{Overall}} \\
            & $\rho$ & $\tau$ & $\rho$ & $\tau$ & $\rho$ & $\tau$ & $\rho$ & $\tau$ \\
        \midrule
            Inter-Annotator & 84.29 & 74.41 & 85.74 & 78.32 & 94.52 & 90.28 & 86.92 & 78.14 \\
        \midrule
            \multicolumn{8}{l}{\textbf{Traditional Methods}} \\
        \midrule
            FID~\cite{heusel2017gans} & -10.75 & -7.56 & -9.65 & -6.88 & -3.54 & -2.70 & -12.31 & -8.62 \\
            LPIPS$_\text{AlexNet}$~\cite{zhang2018perceptual} & -5.46 & -3.86 & -6.83 & -4.54 & -11.68 & -8.63 & -12.44 & -8.56 \\
            DreamSim~\cite{fu2023dreamsim} & -12.48 & -9.04 & -14.81 & -10.83 & -5.56 & -4.14 & -13.82 & -9.68 \\
            CLIPScore~\cite{hessel2021clipscore} & 17.02 & 11.34 & 10.68 & 7.56 & 13.12 & 9.49 & 15.05 & 10.16 \\
            BLIPv2Score~\cite{li2023blip} & 18.97 & 13.04 & 19.67 & 13.57 & 28.07 & 20.89 & 21.52 & 14.23 \\
            PickScore~\cite{kirstain2023pick} & 28.35 & 20.28 & 33.25 & 23.90 & \textbf{39.73} & \textbf{29.63} & 39.44 & 28.03 \\
            ImageReward~\cite{xu2023imagereward} & \textbf{31.29} & \textbf{21.73} & \textbf{43.88} & \textbf{31.74} & 36.43 & 26.96 & \textbf{40.46} & \textbf{28.39} \\
        \midrule
            \multicolumn{8}{l}{\textbf{LLM-based \& MLLM-based Methods}} \\
        \midrule
            LLMScore$_\text{GPT-4}$~\cite{lu2023llmscore} & 18.65 & 13.19 & 29.45 & 21.76 & 21.42 & 16.45 & 30.96 & 22.28 \\
            TIFA$_\text{mPLUG}$~\cite{hu2023tifa} & 21.40 & 15.15 & 39.82 & 29.83 & 22.65 & 17.92 & 32.52 & 24.55 \\
            DSG$_\text{Dependent}$~\cite{Cho2024DSG} & 30.00 & 22.45 & 43.02 & 33.64 & 39.34 & 31.89 & 45.82 & 35.12 \\
            DSG$_\text{Independent}$ & 30.87 & 23.21 & 43.50 & 34.19 & \textbf{44.19} & 33.49 & 47.04 & 36.55 \\
            VQAScore$_\text{CLIP-FlanT5-XXL}$~\cite{vqascore} & 43.59 & 30.35 & \textbf{48.50} & 35.32 & 42.70 & 32.01 & 51.16 & 37.12 \\
            VIEScore$_\text{MiniCPM-V-2.6}$ & 32.25 & 23.63 & 27.42 & 20.68 & 31.75 & 24.82 & 29.41 & 22.50 \\
            VIEScore$_\text{GPT-4o}$~\cite{ku2024viescore} & \textcolor{red}{\textbf{48.60}} & \textbf{35.74} & 46.31 & 34.58 & 43.90 & \textbf{34.16} & 55.45 & 41.70 \\
            T2I-Eval$_\text{GPT-4o}$~\cite{tu2024automatic} & 44.29 & 32.66 & 47.78 & \textbf{35.84} & 39.06 & 31.58 & 55.66 & 42.85 \\
            T2I-Eval$_{\text{MiniCPM-V-2.6}^{*}}$ & 45.03 & 33.82 & 42.87 & 32.88 & 41.72 & 33.92 & \textbf{58.02} & \textcolor{red}{\textbf{44.09}} \\
        \midrule
            \multicolumn{8}{l}{\textbf{Our Method}} \\
        \midrule
            T2I-Eval-R1$_\text{Enhance}$ & \textbf{47.68} & \textcolor{red}{\textbf{36.20}} & \textcolor{red}{\textbf{51.69}} & \textcolor{red}{\textbf{40.72}} & \textcolor{red}{\textbf{51.31}} & \textcolor{red}{\textbf{40.43}} & \textcolor{red}{\textbf{58.74}} & \textbf{43.80} \\
        \bottomrule
        \end{tabular*} 
    }
\end{table*}

\paragraph{T2I-Eval-R1$_\textbf{Enhance}$ almost achieves the strongest in-domain correlations across all three training dimensions}
In the appearance quality, intrinsic and relationship attribute consistency dimensions—exactly those seen during training—our \textbf{Enhance} model variant matches or surpasses both proprietary (GPT-4o–based) and open-source baselines. It is worth noting that our method is sinificantly ahead of previous baselines in intrinsic and relationship attribute consistency, showing great capabilities in learning human preference from simple coarse-grained annotated scores. This validates that T2I-Eval-R1 model produces more accurate interpretable scores when directly aligned to training axes. The exceeded performance of our model compared with GPT-4o-based methods also confirms that T2I evaluation is not limited to proprietary APIs currently.

\paragraph{Sub-optimal performance on the evaluation of Appearance Quality}
While the T2I-Eval-R1 objective demonstrates strong performance across two evaluation dimensions, surpassing prior methods by a large margin, its effectiveness on appearance quality remains slightly inferior to that of the GPT-4o-based VIEScore.Though none of the existing methods using open-source models can bridge the gap, it is still worth exploring whether it is caused by the deficiency of open-source MLLM's visual ability, and whether it can be bridged with innovative optimizing strategies.

\begin{wraptable}{r}{0.4\textwidth}
    \small
    \vspace{-5pt}
    \centering
    \caption{\label{tab:tifa_results} Comparison of previous methods and ours on TIFA v1.0 benchmark, with top scores of each category in \textbf{bold} and top scores of all methods in \textcolor{red}{\textbf{red}}.}
    \resizebox{1.0\linewidth}{!}{
        \begin{tabular*}{1.15\linewidth}{lcc}
        \toprule
            \multirow{2}{*}[-0.0pt]{\textbf{Method}} & \multicolumn{2}{c}{\textbf{Faithfulness}} \\
            & $\rho$ & $\tau$ \\
        \midrule
            Inter-Annotator & 72.22 & 63.85 \\
        \midrule
            CLIPScore~\cite{hessel2021clipscore} & 33.82 & 24.56 \\
            BLIPv2Score~\cite{li2023blip} & 40.49 & 29.44 \\
            PickScore~\cite{kirstain2023pick} & 42.79 & 31.37 \\
            ImageReward~\cite{xu2023imagereward} & \textbf{62.11} & \textbf{46.59} \\
        \midrule
            LLMScore$_\text{GPT-4}$~\cite{lu2023llmscore} & 49.69 & 37.53 \\
            TIFA$_\text{mPLUG}$~\cite{hu2023tifa} & 59.22 & 47.17 \\
            DSG$_\text{Dependent}$~\cite{Cho2024DSG} & 60.46 & 48.93 \\
            DSG$_\text{Independent}$ & 61.08 & 49.54 \\
            VQAScore$_\text{CLIP-FlanT5-XXL}$~\cite{vqascore} & \textbf{69.50} & \textbf{53.21} \\
            VIEScore$_\text{MiniCPM-V-2.6}$ & 37.23 & 27.85 \\
            VIEScore$_\text{GPT-4o}$~\cite{ku2024viescore} & 53.88 & 40.65 \\
            T2I-Eval$_{\text{MiniCPM-V-2.6}^{*}}$ & 60.61 & 46.92 \\
        \midrule
            T2I-Eval-R1$_\text{Base}$ & 69.03 & 53.53 \\
            T2I-Eval-R1$_\text{Hybrid}$ & \textcolor{red}{\textbf{70.43}} & \textcolor{red}{\textbf{55.10}} \\
        \bottomrule
        \end{tabular*}
    }
    \vspace{-10pt}
\end{wraptable}

After verifying the ability of T2I-Eval-R1 objective to fit human preferences in given dimensions, we further conduct evaluation for it on TIFA v1.0 benchmark to validate its generalization ability in the out of domain (OOD) evaluation dimension. Here, apart from the \textbf{Base} variant, we additionally use \textbf{T2I-Eval-R1}$_\textbf{General}$ for evaluation since the inclusion of pairwise data from ImageRewardDB may strengthen the general T2I evaluation abiltiy of the evaluator, preventing the MLLM from overfitting on the given dimensions of T2I-Eval dataset. The results are illustrated in Table~\ref{tab:tifa_results}. Our \textbf{General} variant registers the highest Spearman's $\rho$ (70.43) and Kendall's $\tau$ (55.10), outperforming strong LLM- and MLLM-based methods and slightly surpasses our \textbf{Base} variant. This demonstrates our framework's flexibility: training on coarse scores suffices to adapt the same policy to new semantic criteria, and the inclusion of datasets from various distributions can push the boundary further.

Apart from the single-wise experiments, we also compare our method with several strong baselines in pairwise evaluation on ImageReward. As illustrated in Table~\ref{tab:image_reward}, our \textbf{Pair} variant achieves preference accuracy of 66.07—surpassing prior reward-model and generative chain-of-thought baselines—showing that our confidence-based formulation accurately captures nuanced human preferences than the previous binary approach.

\begin{table}[htbp]
    \centering
    \caption{Comparison of pairwise preference accuracy between our method and baselines on ImageReward benchmark.}
    \resizebox{1.0\linewidth}{!}{
        \begin{tabular}[width=1.0\linewidth]{ccccccc}
            \toprule
                \textbf{Method} & \textbf{CLIPScore} & \textbf{BLIPScore} & \textbf{ImageReward} & \textbf{UnifiedReward} & \textbf{UnifiedReward-Think} & \textbf{T2I-Eval-R1} \\
            \midrule
                Preference Acc. & 54.82 & 57.76 & 65.14 & 55.90 & 64.24 & \textbf{66.07} \\
            \bottomrule
        \end{tabular}
    }
    \label{tab:image_reward}
\end{table}

\vspace{-10pt}
\subsection{Ablation Study on the Rejection Sampling of T2I-Eval-R1$_\textbf{Enhance}$}
\vspace{-10pt}
\begin{table*}[htbp]
    \small
    \centering
    \caption{\label{tab:base_vs_enhance} Comparison on T2I-Eval benchmark of our base variant of T2I-Eval-R1 and the enhanced variant which introduces rejection sampling and further GRPO.}
    \resizebox{0.8\linewidth}{!}{
        \begin{tabular*}{0.85\linewidth}{lcccccccc}
        \toprule
             \multirow{2}{*}[-0.0pt]{\textbf{Method}} & \multicolumn{2}{c}{\textbf{Appearance}} & \multicolumn{2}{c}{\textbf{Intrinsic}} & \multicolumn{2}{c}{\textbf{Relationship}} & \multicolumn{2}{c}{\textbf{Overall}} \\
            & $\rho$ & $\tau$ & $\rho$ & $\tau$ & $\rho$ & $\tau$ & $\rho$ & $\tau$ \\
        \midrule
            T2I-Eval-R1$_\text{Base}$ & 47.07 & 35.15 & \textbf{52.10} & \textbf{40.98} & 50.46 & 39.73 & 58.13 & 42.76 \\
            T2I-Eval-R1$_\text{Enhance}$ & \textbf{47.68} & \textbf{36.20} & 51.69 & 40.72 & \textbf{51.31} & \textbf{40.43} & \textbf{58.74} & \textbf{43.80} \\
        \bottomrule
        \end{tabular*} 
    }
\end{table*}

To validate the effectiveness of the proposed rejection sampling and further GRPO of \textbf{T2I-Eval-R1}$_\textbf{Enhance}$, we compared its performance with the \textbf{Base} variant on T2I-Eval benchmark. As illustrated in Table~\ref{tab:base_vs_enhance}, the exclusion of the enhance method brings slight degradation in the evaluation ability on appearance quality and relationship attribute consistency. Surprisingly, the performance on intrinsic attribute consistency improved slightly, showing that the enhance method does not always brings increment to the evaluator's performance. Generally speaking, the involving of this method still helps us to get the best overall evaluation performance.

\subsection{Ablation Study on Continuous Rewards in Outcome-based Reinforcement Learning}

As described in Section~\ref{subsec:objective}, we propose a novel continuous reward function for GRPO training in both single-wise and pairwise evaluation tasks. To assess the effectiveness of this design, we conduct an ablation study by replacing the continuous reward with a binary reward function, where correct answers receive a reward of 1 and incorrect ones receive 0. All other settings remain unchanged for both \textbf{T2I-Eval-R1}$_\textbf{Base}$ and \textbf{T2I-Eval-R1}$_\textbf{Pair}$.

\begin{table}[htbp]
    \centering
    \caption{\label{tab:reward_ablation_single}Comparison of \textbf{T2I-Eval-R1}$_\text{Base}$ trained with continuous and binary rewards. We report the results of overall quality on T2I-Eval and faithfulness on TIFA v1.0.}
    \resizebox{0.55\linewidth}{!}{
        \begin{tabular*}{0.6\linewidth}{lcccc}
            \toprule
                 \multirow{2}{*}[-0.0pt]{\textbf{Method}} & \multicolumn{2}{c}{\textbf{T2I-Eval}} & \multicolumn{2}{c}{\textbf{TIFA v1.0}} \\
                & $\rho$ & $\tau$ & $\rho$ & $\tau$ \\
            \midrule
                T2I-Eval-R1$_\text{Base}$ & \textbf{58.74} & \textbf{43.80} & \textbf{69.03} & \textbf{53.53} \\
                T2I-Eval-R1$_\text{Base-Binary}$ & 47.25 & 35.56 & 61.03 & 46.92 \\
            \bottomrule
        \end{tabular*}
    }
\end{table}

\begin{table}[htbp]
    \centering
    \caption{\label{tab:reward_ablation_pair}Comparison of \textbf{T2I-Eval-R1}$_\text{Pair}$ trained with continuous and binary rewards. We report preference accuracy (\%) on the ImageReward test set.}
    \resizebox{0.6\linewidth}{!}{
        \begin{tabular}[width=1.0\linewidth]{ccc}
            \toprule
                \textbf{Method} & \textbf{T2I-Eval-R1}$_\textbf{Pair}$ & \textbf{T2I-Eval-R1}$_\textbf{Pair-Binary}$ \\
            \midrule
                Preference Acc. & \textbf{66.07} & 65.19 \\
            \bottomrule
        \end{tabular}
    }
    \vspace{-12pt}
\end{table}

As shown in Table~\ref{tab:reward_ablation_single} and Table~\ref{tab:reward_ablation_pair}, replacing the continuous reward with a binary one leads to a noticeable performance drop across all benchmarks, particularly in the single-wise evaluation tasks. In particular, \textbf{T2I-Eval-R1}$_\textbf{Base-Binary}$ achieves performance comparable to DSG~\cite{Cho2024DSG} on both T2I-Eval and TIFA v1.0, yet significantly underperforms compared to \textbf{T2I-Eval-R1}$_\textbf{Base}$ and other strong baselines.
For pairwise evaluation, the performance gap between continuous and binary rewards is smaller. This may be attributed to the relatively simpler reward structure in pairwise tasks (i.e., three possible choices versus a continuous score spectrum).

Overall, this ablation study demonstrates the effectiveness of the proposed continuous reward function in training MLLM-based evaluators, particularly in enhancing single-wise evaluation performance.

\subsection{Quality of Interpretable Evaluation}
We conducted some complementary studies to assess the quality of the chain-of-thought rationales generated by T2I-Eval-R1, comparing against LLMScore, VIEScore, UnifiedReward-Think, and T2I-Eval evaluators. For single-wise evaluation, we first leveraged GPT-4o as an automatic judge to rate each rationale against the human-annotated gold explanations from the T2I-Eval benchmark. Then, to guard against potential biases of a model-based judge, we performed a human preference study on a sampled subset, measuring direct annotator agreement with GPT-4o's rankings. Both evaluations confirm that T2I-Eval-R1 produces superior, more human-aligned rationales. For pairwise evaluation, since there is no reference rationale in the ImageReward dataset, we only performed human annotation to evaluate the quality of rationales.

\subsubsection{Automatic GPT-4o Assessment}

\begin{wrapfigure}{r}{0.5\textwidth}
    \vspace{-15pt}
    \resizebox{\linewidth}{!}{%
        \includegraphics[width=\textwidth]{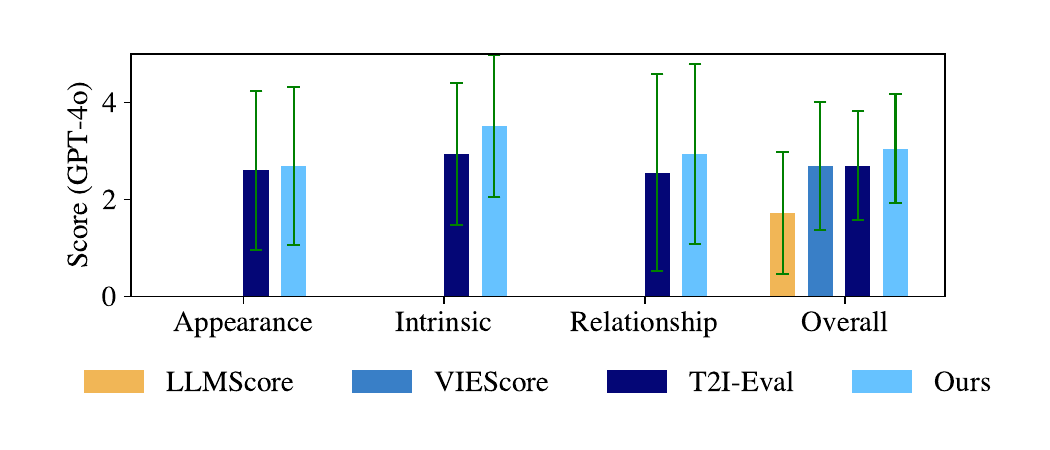}
    }
    \caption{
        Comparison of our method with representative baselines in the results of GPT-4o-based meta-evaluation for interpretable evaluation.
    }
    \label{fig:subjective_eval_gpt-4o}
    \vspace{-13pt}
\end{wrapfigure}

Using the T2I-Eval benchmark, we prompted GPT-4o to score each method's rationale for clarity, completeness, and fidelity to the reference explanation. The results are illustrated in Figure~\ref{fig:subjective_eval_gpt-4o}.

\paragraph{Overall wins}: T2I-Eval-R1 achieved the highest average quality score in the Overall evaluation, outperforming LLMScore, VIEScore, and the T2I-Eval evaluators.
\paragraph{Dimension-wise superiority}: Across each of the three dimensions—appearance quality, intrinsic and relationship attribute consistency—T2I-Eval-R1's rationales consistently received higher ratings than T2I-Eval, indicating that out method even outperforms naive supervised fine-tuning with an larger amount of fine-grained training data in the same distribution of the reference rationales.

\paragraph{Quality stability}: Among all the comparisons in the interpretable evaluation, our method consistently has the lowest standard deviation of the scores, revealing that the quality of interpretable rationales in our method is stabler than previous methods.

While GPT-4o provides a scalable, fine-grained critique of rationale text, its judgments may be subject to its own reasoning biases and prompt sensitivity. To ensure these automatic scores reflect real human preferences, we complemented this analysis with a direct human study.

\subsubsection{Human Preference Study}

\begin{table}[htbp]
    \centering
    \caption{Quality of subjective evaluation in our method compared with baseline methods.}
    \resizebox{1.0\linewidth}{!}{
        \begin{tabular}[width=1.0\linewidth]{cccccccccccccc}
            \toprule
                \multirow{3}{*}[-2.0pt]{\textbf{Category}} & \multicolumn{6}{c}{\textbf{Overall Evaluation}} & \multicolumn{6}{c}{\textbf{Single Dimension (T2I-Eval)}} & \textbf{Pairwise} \\
            \cmidrule{2-14}
                & \multicolumn{2}{c}{LLMScore} & \multicolumn{2}{c}{VIEScore} & \multicolumn{2}{c}{T2I-Eval} & \multicolumn{2}{c}{Appearance} & \multicolumn{2}{c}{Intrinsic} & \multicolumn{2}{c}{Relationship} & \multirow{2}{*}{\begin{tabular}[c]{@{}c@{}}UnifiedReward\\-Think\end{tabular}} \\
                & -T & +T & -T & +T & -T & +T & -T & +T & -T & +T & -T & +T & \\
            \midrule
                Human Preference & 62.0 & 94.0 & 14.0 & 79.0 & 40.0 & 89.0 & 46.0 & 78.0 & 60.0 & 84.0 & 34.0 & 94.0 & 55.0 \\
                GPT-4o Preference & 77.1 & 84.4 & 51.2 & 63.1 & 56.1 & 66.1 & 34.2 & 70.1 & 49.5 & 73.8 & 40.5 & 73.4 & - \\
                GPT-4o Accuracy & \multicolumn{2}{c}{82.4} & \multicolumn{2}{c}{58.6} & \multicolumn{2}{c}{63.7} & \multicolumn{2}{c}{55.9} & \multicolumn{2}{c}{63.2} & \multicolumn{2}{c}{80.0} & - \\
            \bottomrule
        \end{tabular}
    }
    \label{tab:human_annotation}
\end{table}

We conduct human annotation for the quality of rationale text on both single-wise and pairwise settings. For single-wise evaluations, we randomly sampled 200 overall evaluation examples in comparison with each baseline, and 50 examples for each dimension in comparison with T2I-Eval. Annotators were asked which explanation they preferred in terms of interpretability and usefulness~\cite{xu2025decider}. We then measured:
\begin{itemize}[nosep, leftmargin=25pt]
    \item \textbf{Preference accuracy of GPT-4o}: the fraction of cases where GPT-4o’s winner matched the human judgement.
    \item \textbf{Human preference rate}: the percentage of samples where annotators favored T2I-Eval-R1 over the competing method (including or excluding ties).
\end{itemize}
The results are illustrated in Table~\ref{tab:human_annotation}. From these results, we can conclude that
(1) High human agreement (averaged 67.3\%) with GPT-4o's rankings, validating GPT-4o as a reliable proxy judge for rationale quality;
(2) Clear human preference (with tie) for T2I-Eval-R1's explanations in over 75\% of comparisons, demonstrating that its CoT is not only more precise according to an LLM judge but also genuinely more helpful to human readers.
For pairwise evaluation, we compared the rationale from our method and UnifiedReward-Think. The win rate of our method is 55\%, indicating better pairwise reasoning capabilities of T2I-Eval-R1 series models.

\section{Conclusion}
We have presented T2I-Eval-R1, a framework that trains open-source MLLMs to produce both scalar scores and CoT rationales for T2I evaluation using only coarse‐grained quality labels.  By integrating GRPO with continuous reward functions, the optimized models yield interpretable evaluations that align more closely with human judgments than prior approaches across diverse dimensions. Extensive experiments demonstrate that T2I-Eval-R1 not only matches or exceeds GPT-4o-based baselines in correlations with human judgements, but also generates better explanations, as confirmed by both automatic evaluation and human annotation.  Importantly, it achieves these gains without reliance on fine-grained annotations, offering a scalable and generalizable solution for evaluating T2I generative models. By lowering the barrier to interpretable automatic evaluation, T2I-Eval-R1 paves the way for more rigorous, transparent evaluation of generative systems in both research and deployment.

\bibliographystyle{ref}  
\small
\bibliography{reference}

\begin{thebibliography}{39}
\providecommand{\natexlab}[1]{#1}
\providecommand{\url}[1]{\texttt{#1}}
\expandafter\ifx\csname urlstyle\endcsname\relax
  \providecommand{\doi}[1]{doi: #1}\else
  \providecommand{\doi}{doi: \begingroup \urlstyle{rm}\Url}\fi

\bibitem[Achiam et~al.(2023)Achiam, Adler, Agarwal, Ahmad, Akkaya, Aleman, Almeida, Altenschmidt, Altman, Anadkat, et~al.]{achiam2023gpt}
Achiam , J., Adler , S., Agarwal , S., Ahmad , L., Akkaya , I., Aleman , F.~L., Almeida , D., Altenschmidt , J., Altman , S., Anadkat , S., \& others  (2023)
\newblock Gpt-4 technical report.
\newblock \emph{arXiv preprint arXiv:2303.08774}

\bibitem[Anil et~al.(2023)Anil, Borgeaud, Wu, Alayrac, Yu, Soricut, Schalkwyk, Dai, Hauth, Millican, Silver, Petrov, Johnson, Antonoglou, Schrittwieser, Glaese, Chen, Pitler, Lillicrap, Lazaridou, Firat, Molloy, Isard, Barham, Hennigan, Lee, Viola, Reynolds, Xu, Doherty, Collins, Meyer, Rutherford, Moreira, Ayoub, Goel, Tucker, Piqueras, Krikun, Barr, Savinov, Danihelka, Roelofs, White, Andreassen, von Glehn, Yagati, Kazemi, Gonzalez, Khalman, Sygnowski, and et~al.]{gemini}
Anil , R., Borgeaud , S., Wu~, Y., Alayrac , J., Yu~, J., Soricut , R., Schalkwyk , J., Dai , A.~M., Hauth , A., Millican , K., Silver , D., Petrov , S., Johnson , M., Antonoglou , I., Schrittwieser , J., Glaese , A., Chen , J., Pitler , E., Lillicrap , T.~P., Lazaridou , A., Firat , O., Molloy , J., Isard , M., Barham , P.~R., Hennigan , T., Lee , B., Viola , F., Reynolds , M., Xu~, Y., Doherty , R., Collins , E., Meyer , C., Rutherford , E., Moreira , E., Ayoub , K., Goel , M., Tucker , G., Piqueras , E., Krikun , M., Barr , I., Savinov , N., Danihelka , I., Roelofs , B., White , A., Andreassen , A., Glehn , T., Yagati , L., Kazemi , M., Gonzalez , L., Khalman , M., Sygnowski , J., \& al.  (2023)
\newblock Gemini: {A} family of highly capable multimodal models.
\newblock \emph{CoRR} {\bfseries abs/2312.11805}.

\bibitem[Bai et~al.(2025)Bai, Chen, Liu, Wang, Ge, Song, Dang, Wang, Wang, Tang, et~al.]{bai2025qwen2}
Bai , S., Chen , K., Liu , X., Wang , J., Ge~, W., Song , S., Dang , K., Wang , P., Wang , S., Tang , J., \& others  (2025)
\newblock Qwen2. 5-vl technical report.
\newblock \emph{arXiv preprint arXiv:2502.13923}

\bibitem[Bradley and Terry(1952)]{bradley1952rank}
Bradley , R.~A. \& Terry , M.~E. (1952)
\newblock Rank analysis of incomplete block designs: I. the method of paired comparisons.
\newblock \emph{Biometrika} {\bfseries 39}\penalty0 (3/4):\penalty0 324--345.

\bibitem[Cho et~al.(2024)Cho, Hu, Baldridge, Garg, Anderson, Krishna, Bansal, Pont-Tuset, and Wang]{Cho2024DSG}
Cho , J., Hu~, Y., Baldridge , J., Garg , R., Anderson , P., Krishna , R., Bansal , M., Pont-Tuset , J., \& Wang , S. (2024)
\newblock Davidsonian scene graph: Improving reliability in fine-grained evaluation for text-to-image generation. In
\newblock \emph{ICLR}

\bibitem[Dao(2023)]{dao2023flashattention}
Dao , T. (2023)
\newblock Flashattention-2: Faster attention with better parallelism and work partitioning.
\newblock \emph{arXiv preprint arXiv:2307.08691}

\bibitem[Eisenstein et~al.(2023)Eisenstein, Nagpal, Agarwal, Beirami, D'Amour, Dvijotham, Fisch, Heller, Pfohl, Ramachandran, et~al.]{eisenstein2023helping}
Eisenstein , J., Nagpal , C., Agarwal , A., Beirami , A., D'Amour , A., Dvijotham , D., Fisch , A., Heller , K., Pfohl , S., Ramachandran , D., \& others  (2023)
\newblock Helping or herding? reward model ensembles mitigate but do not eliminate reward hacking.
\newblock \emph{arXiv preprint arXiv:2312.09244}

\bibitem[Esser et~al.(2024)Esser, Kulal, Blattmann, Entezari, M{\"u}ller, Saini, Levi, Lorenz, Sauer, Boesel, et~al.]{esser2024scaling}
Esser , P., Kulal , S., Blattmann , A., Entezari , R., M{\"u}ller , J., Saini , H., Levi , Y., Lorenz , D., Sauer , A., Boesel , F., \& others  (2024)
\newblock Scaling rectified flow transformers for high-resolution image synthesis. In
\newblock \emph{Forty-first international conference on machine learning}

\bibitem[Fu et~al.(2023)Fu, Tamir, Sundaram, Chai, Zhang, Dekel, and Isola]{fu2023dreamsim}
Fu~, S., Tamir , N., Sundaram , S., Chai , L., Zhang , R., Dekel , T., \& Isola , P. (2023)
\newblock Dreamsim: Learning new dimensions of human visual similarity using synthetic data.
\newblock \emph{Advances in Neural Information Processing Systems} {\bfseries 36}:\penalty0 50742--50768.

\bibitem[Guo et~al.(2025)Guo, Yang, Zhang, Song, Zhang, Xu, Zhu, Ma, Wang, Bi, et~al.]{guo2025deepseek}
Guo , D., Yang , D., Zhang , H., Song , J., Zhang , R., Xu~, R., Zhu , Q., Ma~, S., Wang , P., Bi~, X., \& others  (2025)
\newblock Deepseek-r1: Incentivizing reasoning capability in llms via reinforcement learning.
\newblock \emph{arXiv preprint arXiv:2501.12948}

\bibitem[Gupta et~al.(2025)Gupta, Ahuja, Lin, Roy, Oosterhuis, de~Rijke, and Shukla]{gupta2025simple}
Gupta , S., Ahuja , C., Lin , T.-Y., Roy , S.~D., Oosterhuis , H., Rijke , M., \& Shukla , S.~N. (2025)
\newblock A simple and effective reinforcement learning method for text-to-image diffusion fine-tuning.
\newblock \emph{arXiv preprint arXiv:2503.00897}

\bibitem[Hara et~al.(2018)Hara, Adams, Milland, Savage, Callison-Burch, and Bigham]{hara2018data}
Hara , K., Adams , A., Milland , K., Savage , S., Callison-Burch , C., \& Bigham , J.~P. (2018)
\newblock A data-driven analysis of workers' earnings on amazon mechanical turk. In
\newblock \emph{Proceedings of the 2018 CHI conference on human factors in computing systems}
\newblock pages 1--14.

\bibitem[Hessel et~al.(2021)Hessel, Holtzman, Forbes, Bras, and Choi]{hessel2021clipscore}
Hessel , J., Holtzman , A., Forbes , M., Bras , R.~L., \& Choi , Y. (2021)
\newblock Clipscore: A reference-free evaluation metric for image captioning.
\newblock \emph{arXiv preprint arXiv:2104.08718}

\bibitem[Heusel et~al.(2017)Heusel, Ramsauer, Unterthiner, Nessler, and Hochreiter]{heusel2017gans}
Heusel , M., Ramsauer , H., Unterthiner , T., Nessler , B., \& Hochreiter , S. (2017)
\newblock Gans trained by a two time-scale update rule converge to a local nash equilibrium.
\newblock \emph{Advances in neural information processing systems} {\bfseries 30}.

\bibitem[Hu et~al.(2022)Hu, Shen, Wallis, Allen-Zhu, Li, Wang, Wang, Chen, et~al.]{hu2022lora}
Hu~, E.~J., Shen , Y., Wallis , P., Allen-Zhu , Z., Li~, Y., Wang , S., Wang , L., Chen , W., \& others  (2022)
\newblock Lora: Low-rank adaptation of large language models.
\newblock \emph{ICLR} {\bfseries 1}\penalty0 (2):\penalty0 3.

\bibitem[Hu et~al.(2023)Hu, Liu, Kasai, Wang, Ostendorf, Krishna, and Smith]{hu2023tifa}
Hu~, Y., Liu , B., Kasai , J., Wang , Y., Ostendorf , M., Krishna , R., \& Smith , N.~A. (2023)
\newblock Tifa: Accurate and interpretable text-to-image faithfulness evaluation with question answering. In
\newblock \emph{Proceedings of the IEEE/CVF International Conference on Computer Vision}
\newblock pages 20406--20417.

\bibitem[Huang et~al.(2025)Huang, Jia, Zhai, Cao, Ye, Zhao, Xu, Hu, and Lin]{huang2025vision}
Huang , W., Jia , B., Zhai , Z., Cao , S., Ye~, Z., Zhao , F., Xu~, Z., Hu~, Y., \& Lin , S. (2025)
\newblock Vision-r1: Incentivizing reasoning capability in multimodal large language models.
\newblock \emph{arXiv preprint arXiv:2503.06749}

\bibitem[Jin et~al.(2025)Jin, Zeng, Yue, Yoon, Arik, Wang, Zamani, and Han]{jin2025search}
Jin , B., Zeng , H., Yue , Z., Yoon , J., Arik , S., Wang , D., Zamani , H., \& Han , J. (2025)
\newblock Search-r1: Training llms to reason and leverage search engines with reinforcement learning.
\newblock \emph{arXiv preprint arXiv:2503.09516}

\bibitem[Kirstain et~al.(2023)Kirstain, Polyak, Singer, Matiana, Penna, and Levy]{kirstain2023pick}
Kirstain , Y., Polyak , A., Singer , U., Matiana , S., Penna , J., \& Levy , O. (2023)
\newblock Pick-a-pic: An open dataset of user preferences for text-to-image generation.
\newblock \emph{Advances in Neural Information Processing Systems} {\bfseries 36}:\penalty0 36652--36663.

\bibitem[Ku et~al.(2024)Ku, Jiang, Wei, Yue, and Chen]{ku2024viescore}
Ku~, M., Jiang , D., Wei , C., Yue , X., \& Chen , W. (2024)
\newblock Viescore: Towards explainable metrics for conditional image synthesis evaluation. In
\newblock \emph{Proceedings of the 62nd Annual Meeting of the Association for Computational Linguistics (Volume 1: Long Papers)}
\newblock pages 12268--12290.

\bibitem[Li et~al.(2023)Li, Li, Savarese, and Hoi]{li2023blip}
Li~, J., Li~, D., Savarese , S., \& Hoi , S. (2023)
\newblock Blip-2: Bootstrapping language-image pre-training with frozen image encoders and large language models. In
\newblock \emph{International conference on machine learning}
\newblock pages 19730--19742. PMLR.

\bibitem[Lin et~al.(2024)Lin, Pathak, Li, Li, Xia, Neubig, Zhang, and Ramanan]{vqascore}
Lin , Z., Pathak , D., Li~, B., Li~, J., Xia , X., Neubig , G., Zhang , P., \& Ramanan , D. (2024)
\newblock Evaluating text-to-visual generation with image-to-text generation. In
\newblock A.~Leonardis, E.~Ricci, S.~Roth, O.~Russakovsky, T.~Sattler, and G.~Varol, (eds.),
\newblock \emph{Computer Vision - {ECCV} 2024 - 18th European Conference, Milan, Italy, September 29-October 4, 2024, Proceedings, Part {IX}}
\newblock \emph{15067}, pp. \penalty0 366--384. Springer.

\bibitem[Lu et~al.(2023)Lu, Yang, Li, Wang, and Wang]{lu2023llmscore}
Lu~, Y., Yang , X., Li~, X., Wang , X.~E., \& Wang , W.~Y. (2023)
\newblock Llmscore: Unveiling the power of large language models in text-to-image synthesis evaluation.
\newblock \emph{Advances in Neural Information Processing Systems} {\bfseries 36}:\penalty0 23075--23093.

\bibitem[Peebles and Xie(2023)]{peebles2023scalable}
Peebles , W. \& Xie , S. (2023)
\newblock Scalable diffusion models with transformers. In
\newblock \emph{Proceedings of the IEEE/CVF international conference on computer vision}
\newblock pages 4195--4205.

\bibitem[Podell et~al.(2023)Podell, English, Lacey, Blattmann, Dockhorn, M{\"u}ller, Penna, and Rombach]{podellsdxl}
Podell , D., English , Z., Lacey , K., Blattmann , A., Dockhorn , T., M{\"u}ller , J., Penna , J., \& Rombach , R. (2023)
\newblock Sdxl: Improving latent diffusion models for high-resolution image synthesis. In
\newblock \emph{The Twelfth International Conference on Learning Representations}

\bibitem[Rafailov et~al.(2023)Rafailov, Sharma, Mitchell, Manning, Ermon, and Finn]{rafailov2023direct}
Rafailov , R., Sharma , A., Mitchell , E., Manning , C.~D., Ermon , S., \& Finn , C. (2023)
\newblock Direct preference optimization: Your language model is secretly a reward model.
\newblock \emph{Advances in Neural Information Processing Systems} {\bfseries 36}:\penalty0 53728--53741.

\bibitem[Rajbhandari et~al.(2020)Rajbhandari, Rasley, Ruwase, and He]{rajbhandari2020zero}
Rajbhandari , S., Rasley , J., Ruwase , O., \& He~, Y. (2020)
\newblock Zero: Memory optimizations toward training trillion parameter models. In
\newblock \emph{SC20: International Conference for High Performance Computing, Networking, Storage and Analysis}
\newblock pages 1--16. IEEE.

\bibitem[Rombach et~al.(2022)Rombach, Blattmann, Lorenz, Esser, and Ommer]{rombach2022high}
Rombach , R., Blattmann , A., Lorenz , D., Esser , P., \& Ommer , B. (2022)
\newblock High-resolution image synthesis with latent diffusion models. In
\newblock \emph{Proceedings of the IEEE/CVF conference on computer vision and pattern recognition}
\newblock pages 10684--10695.

\bibitem[Salimans et~al.(2016)Salimans, Goodfellow, Zaremba, Cheung, Radford, and Chen]{salimans2016improved}
Salimans , T., Goodfellow , I., Zaremba , W., Cheung , V., Radford , A., \& Chen , X. (2016)
\newblock Improved techniques for training gans.
\newblock \emph{Advances in neural information processing systems} {\bfseries 29}.

\bibitem[Touvron et~al.(2023)Touvron, Martin, Stone, Albert, Almahairi, Babaei, Bashlykov, Batra, Bhargava, Bhosale, et~al.]{touvron2023llama}
Touvron , H., Martin , L., Stone , K., Albert , P., Almahairi , A., Babaei , Y., Bashlykov , N., Batra , S., Bhargava , P., Bhosale , S., \& others  (2023)
\newblock Llama 2: Open foundation and fine-tuned chat models.
\newblock \emph{arXiv preprint arXiv:2307.09288}

\bibitem[Tu et~al.(2024)Tu, Ma, Lan, Zhao, Huang, and Mao]{tu2024automatic}
Tu~, R.-C., Ma~, Z.-A., Lan , T., Zhao , Y., Huang , H., \& Mao , X.-L. (2024)
\newblock Automatic evaluation for text-to-image generation: Task-decomposed framework, distilled training, and meta-evaluation benchmark.
\newblock \emph{arXiv preprint arXiv:2411.15488}

\bibitem[Wang et~al.(2025{\natexlab{a}})Wang, Li, Zang, Wang, Lu, Jin, and Wang]{wang2025unifiedmultimodal}
Wang , Y., Li~, Z., Zang , Y., Wang , C., Lu~, Q., Jin , C., \& Wang , J. (2025.
\newblock {\natexlab{a}})
\newblock Unified multimodal chain-of-thought reward model through reinforcement fine-tuning.
\newblock \emph{arXiv preprint arXiv:2505.03318}

\bibitem[Wang et~al.(2025{\natexlab{b}})Wang, Zang, Li, Jin, and Wang]{wang2025unified}
Wang , Y., Zang , Y., Li~, H., Jin , C., \& Wang , J. (2025.
\newblock {\natexlab{b}})
\newblock Unified reward model for multimodal understanding and generation.
\newblock \emph{arXiv preprint arXiv:2503.05236}

\bibitem[Welleck et~al.(2020)Welleck, Kulikov, Roller, Dinan, Cho, and Weston]{welleck2020neural}
Welleck , S., Kulikov , I., Roller , S., Dinan , E., Cho , K., \& Weston , J. (2020)
\newblock Neural text degeneration with unlikelihood training. In
\newblock \emph{8th International Conference on Learning Representations, ICLR 2020}

\bibitem[Xu et~al.(2025)Xu, Lan, Ji, Yu, Wang, Gao, Dong, Qian, Li, Bi, et~al.]{xu2025decider}
Xu~, C., Lan , T., Ji~, Y., Yu~, C., Wang , W., Gao , J., Dong , Q., Qian , K., Li~, P., Bi~, W., \& others  (2025)
\newblock Decider: A dual-system rule-controllable decoding framework for language generation.
\newblock \emph{IEEE Transactions on Knowledge and Data Engineering}

\bibitem[Xu et~al.(2023)Xu, Liu, Wu, Tong, Li, Ding, Tang, and Dong]{xu2023imagereward}
Xu~, J., Liu , X., Wu~, Y., Tong , Y., Li~, Q., Ding , M., Tang , J., \& Dong , Y. (2023)
\newblock Imagereward: Learning and evaluating human preferences for text-to-image generation.
\newblock \emph{Advances in Neural Information Processing Systems} {\bfseries 36}:\penalty0 15903--15935.

\bibitem[Zhang et~al.(2025)Zhang, Huang, Yao, Liu, Zhang, Lu, and Tao]{zhang2025r1}
Zhang , J., Huang , J., Yao , H., Liu , S., Zhang , X., Lu~, S., \& Tao , D. (2025)
\newblock R1-vl: Learning to reason with multimodal large language models via step-wise group relative policy optimization.
\newblock \emph{arXiv preprint arXiv:2503.12937}

\bibitem[Zhang et~al.(2018)Zhang, Isola, Efros, Shechtman, and Wang]{zhang2018perceptual}
Zhang , R., Isola , P., Efros , A.~A., Shechtman , E., \& Wang , O. (2018)
\newblock The unreasonable effectiveness of deep features as a perceptual metric. In
\newblock \emph{CVPR}

\bibitem[Zhao et~al.(2025)Zhao, Huang, Hu, Wang, Mao, Zhang, Jiang, Wu, Ai, Wang, et~al.]{zhao2025swift}
Zhao , Y., Huang , J., Hu~, J., Wang , X., Mao , Y., Zhang , D., Jiang , Z., Wu~, Z., Ai~, B., Wang , A., \& others  (2025)
\newblock Swift: a scalable lightweight infrastructure for fine-tuning. In
\newblock \emph{Proceedings of the AAAI Conference on Artificial Intelligence}
\newblock \emph{39}, pp. \penalty0 29733--29735.

\end{thebibliography}
\normalsize

\clearpage
\appendix
\section{Limitations~\label{app:limitations}}

Despite its strengths, T2I-Eval-R1 has the following limitations:

\paragraph{Evaluation across unseen dimensions is constrained}
Although we demonstrate generalization on the faithfulness dimension from TIFA v1.0, the lack of diverse, high-quality human-annotated benchmarks for other aspects (e.g., aesthetic style, composition rules, cultural context) limits our ability to comprehensively validate robustness across truly novel evaluation criteria.

\paragraph{Model and compute resource scope}
All experiments in this work employ the Qwen2.5-VL-7B-Instruct backbone due to our restricted computational budget. Consequently, we have not verified T2I-Eval-R1's effectiveness on other multimodal architectures or larger-scale models, leaving open questions about scaling behavior and cross-model transfer.

\section{Ethical Considerations\label{app:ethical}}
Most of the text-to-image task inputs in our training dataset are sourced from publicly available datasets, ensuring that they pose no harm to individuals or groups. Furthermore, the text generated by multi-modal language models (MLLMs) is carefully curated and processed by human annotators to safeguard privacy and confidentiality. No personally identifiable information (PII) is included. However, it is important to note that the generated images from TIFA-v1.0~\cite{hu2023tifa}, ImageRewardDB~\cite{xu2023imagereward} and T2I-Eval~\cite{tu2024automatic} may contain harmful content. Despite these potential risks, it is crucial to disclose the full scope of this research, as materials from these datasets have been extensively used in safety research within the community. All annotators are compensated fairly, with an hourly wage of approximately \$5.15 USD, which exceeds the average hourly wage of \$3.13 USD on Amazon Mechanical Turk~\cite{hara2018data}.

\section{Benchmark Datasets\label{app:benchmark}}
We assess our models trained with Eval-GRPO on three public benchmarks, each targeting a distinct evaluation mode:
\begin{itemize}[nosep, leftmargin=25pt]
    \item \textbf{T2I-Eval}~\cite{tu2024automatic} (under MIT License): Contains coarse human annotated scores on three fine-grained dimensions—appearance quality, intrinsic attribute consistency, and relationship attribute consistency. We train on its training split and report Spearman's $\rho$ and Kendall's $\tau$ against its held-out test set to verify in-domain performance.
    \item \textbf{TIFA v1.0}~\cite{hu2023tifa} (under Apache 2.0 License): Focuses on text-to-image faithfulness, a dimension unseen during our training. We run T2I-Eval-R1 in single-wise mode on its test prompts and compute Spearman's $\rho$ and Kendall's $\tau$ against the published QA-based faithfulness judgments to measure generalization.
    \item \textbf{ImageReward}~\cite{xu2023imagereward} (under Apache 2.0 License): Provides large-scale pairwise human preferences. We evaluate our pairwise confidence outputs against its ground-truth comparisons, again reporting preference accuracy, to demonstrate our model’s ranking capability.
\end{itemize}

\section{Baseline Methods~\label{app:baseline_methods}}
To contextualize our results, we compare our method against four categories of existing evaluators:
\begin{itemize}[nosep, leftmargin=25pt]
    \item \textbf{Traditional Metrics} (e.g., FID~\cite{heusel2017gans}, Inception Score~\cite{salimans2016improved}): Rely on distributional statistics of deep‐network embeddings to measure overall realism and diversity, but they lack fine-grained, prompt-conditioned assessments.
    \item \textbf{Embedding-based and Learning-based Methods} (e.g., CLIPScore~\cite{hessel2021clipscore}, BLIPv2Score~\cite{li2023blip}, LPIPS~\cite{zhang2018perceptual}, DreamSim~\cite{fu2023dreamsim}, PickScore~\cite{kirstain2023pick}, ImageReward~\cite{xu2023imagereward}): Leverage pretrained vision–language or perceptual models to quantify semantic alignment or visual similarity, offering stronger prompt sensitivity than purely statistical metrics.
    \item \textbf{LLM-based Evaluators} (e.g., LLMScore~\cite{lu2023llmscore}, TIFA~\cite{hu2023tifa}): Use large language models—sometimes combined with visual question answering—to decompose evaluation into subquestions, yielding interpretable scores but often at high computational cost.
    \item \textbf{MLLM-based Chain-of-Thought Methods} (e.g., VIEScore~\cite{ku2024viescore}, T2I-Eval~\cite{tu2024automatic}, UnifiedReward-Think~\cite{wang2025unifiedmultimodal}): Instruct multimodal LLMs to generate detailed rationales alongside ratings, achieving state-of-the-art correlations with human judgments while highlighting the gap between proprietary APIs and open-source alternatives.
\end{itemize}
We select these baselines to cover the spectrum from lightweight, reference-free metrics through sophisticated, rationale-driven evaluators, ensuring a comprehensive assessment of both our in-domain performance and generalization capabilities.

\section{Training Settings~\label{app:training_settings}}
We optimize the open-source MLLM Qwen2.5-VL-7B-Instruct~\cite{bai2025qwen2} to serve as a reasoning-based automatic evaluator. To ensure the optimized model effectively captures the comprehensive information embedded in the training corpus, we set the context length to 4,096 tokens during training, accommodating the majority of samples within the dataset. To optimize the computational efficiency and uphold the performance of the fine-tuned model, we employed Low-Rank Adaptation (LoRA)~\cite{hu2022lora} with the rank of 256 and $\alpha$ of 512. Apart from that, we adopt various methods to accelerate training including ZeRO~\cite{rajbhandari2020zero} and Flash Attention 2~\cite{dao2023flashattention}. The model training was conducted on 4 Nvidia A100-SXM4-80GB GPUs with a group size of 8 over a single epoch. All training is implemented with the SWIFT framework~\cite{zhao2025swift}.

\section{Full Prompt Templates for Involved Settings~\label{app:prompt_template}}

\begin{figure}[h]
\centering
    \includegraphics[width=1.0\linewidth]{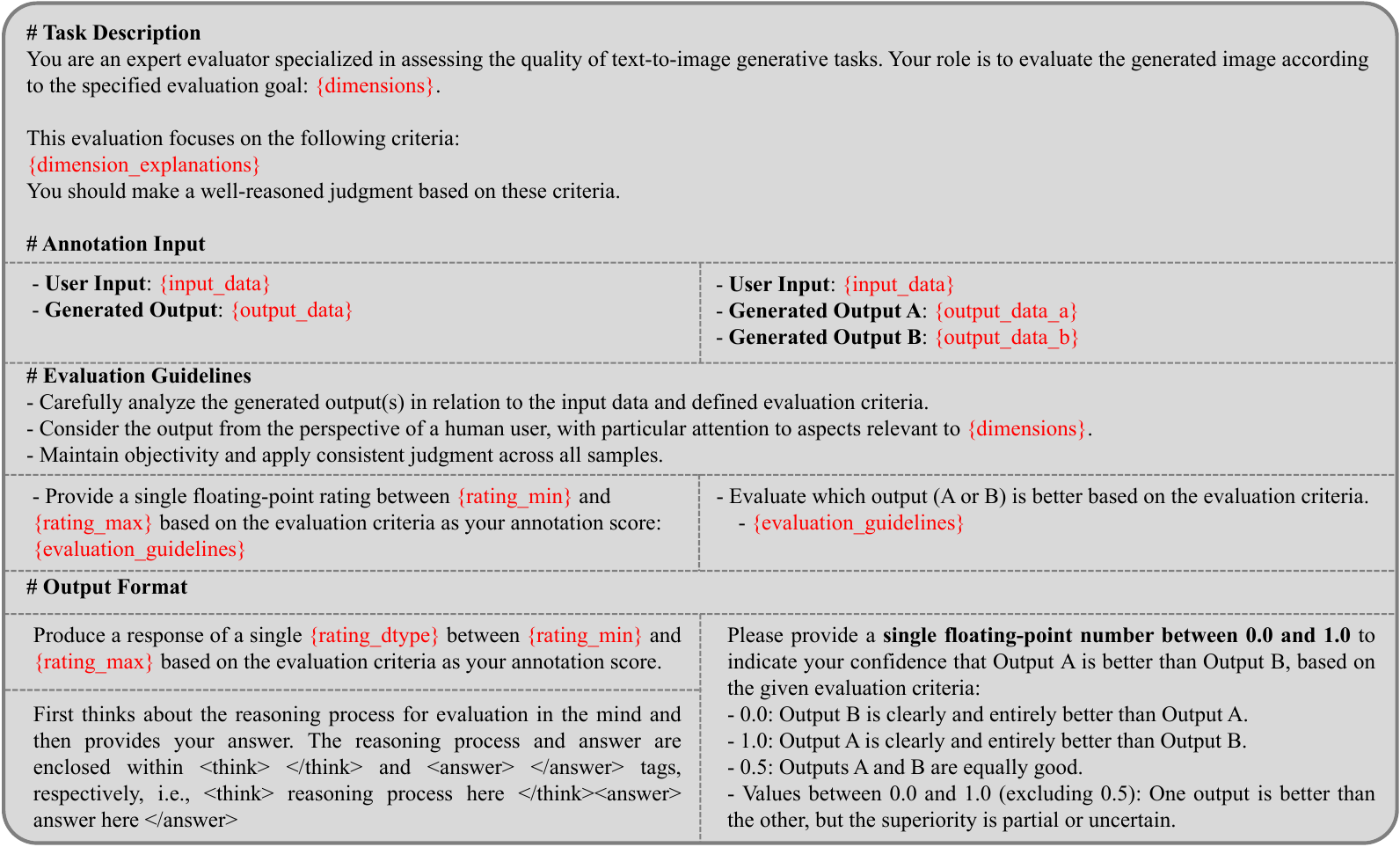}
\caption{Prompt template for T2I-Eval-R1 evaluators. The evaluation prompt is dynamically assembled with this template according to requirements of the specific task. The adjustable parts for single-wise and pairwise protocols are placed on the left and right side, respectively.}
\label{fig:eval_template}
\end{figure}

As described in Section~\ref{subsec:prompt_design}, the four-block prompt assembly framework is illustrated in Figure~\ref{fig:eval_template}. There are various evaluation settings involved in our experiment. Though sharing the same framework, each setting has diverse evaluation dimensions, definitions and guidelines. The settings include:
\begin{itemize}
    \item \textbf{Single-wise}:
        \begin{itemize}
            \item \textbf{Appearance Quality}: As illustrated in Figure~\ref{fig:prompt_appearance};
            \item \textbf{Intrinsic Attribute Consistency}: As illustrated in Figure~\ref{fig:prompt_intrinsic};
            \item \textbf{Relationship Attribute Consistency}: As illustrated in Figure~\ref{fig:prompt_relationship};
            \item \textbf{Overall Quality}: As illustrated in Figure~\ref{fig:prompt_overall};
            \item \textbf{Text-to-Image Faithfulness}: As illustrated in Figure~\ref{fig:prompt_tifa};
        \end{itemize}
    \item \textbf{Pairwise}: As illustrated in Figure~\ref{fig:prompt_imagereward}.
\end{itemize}

\begin{figure}[h]
\scriptsize
\centering
\begin{tcolorbox}
\textbf{\# Task Description}\\
You are an expert evaluator specialized in assessing the quality of text-to-image generative tasks. Your role is to evaluate the generated image according to the specified evaluation goal: \textbf{appearance quality}.\\
\\
This evaluation focuses on the following criteria:\\
* What we mean by ``appearance quality'' is that the appearance of the entity in the generated image should be realistic, aesthetically pleasing, and aligns with human intuition.  \\
You should make a well-reasoned judgment based on these criteria.\\
\\
\textbf{\# Annotation Input}\\
- \textbf{User Input}: \textcolor{red}{<text>}\\
- \textbf{Generated Output}: \textcolor{red}{<image>}\\
\\
\textbf{\# Evaluation Guidelines}\\
- Carefully analyze the generated output(s) in relation to the input data and defined evaluation criteria.\\
- Consider the output from the perspective of a human user, with particular attention to aspects relevant to \textbf{appearance quality}.\\
- Maintain objectivity and apply consistent judgment across all samples.\\
- Provide a \textbf{single floating-point number between 0.0 and 10.0} based on the evaluation criteria as your annotation score:\\
Evaluate the Appearance Quality of the generated image:\\
\text{\quad}- Determine whether the entity from the user-provided text is in the generated image. If yes, proceed to the next step. If no, give a score of 0.\\
\text{\quad}- Give a score from 0.0 to 10.0:\\
\text{\quad}\text{\quad}- 0-3: The appearance is not realistic, aesthetically pleasing, or align with human intuition at all.\\
\text{\quad}\text{\quad}- 4-7: The appearance is somewhat realistic, aesthetically pleasing, or align with human intuition.\\
\text{\quad}\text{\quad}- 8-10: The appearance is very realistic, aesthetically pleasing, or align with human intuition.\\
\\
\textbf{\# Output Format}\\
Produce a response of a \textbf{single floating-point number between 0.0 and 10.0} based on the evaluation criteria as your annotation score.\\
\end{tcolorbox}
\caption{Prompt template for Appearance Quality evaluation from T2I-Eval dataset.}
\label{fig:prompt_appearance}
\end{figure}

\begin{figure}[h]
\scriptsize
\centering
\begin{tcolorbox}
\textbf{\# Task Description}\\
You are an expert evaluator specialized in assessing the quality of text-to-image generative tasks. Your role is to evaluate the generated image according to the specified evaluation goal: \textbf{intrinsic attribute consistency}.\\
\\
This evaluation focuses on the following criteria:\\
* What we mean by ``intrinsic attribute'' is that the attributes are properties of the entity explicitly mentioned in the input text, such as color, size, shape, material, quantity, etc.  \\
You should make a well-reasoned judgment based on these criteria.\\
\\
\textbf{\# Annotation Input}\\
- \textbf{User Input}: \textcolor{red}{<text>}\\
- \textbf{Generated Output}: \textcolor{red}{<image>}\\
\\
\textbf{\# Evaluation Guidelines}\\
- Carefully analyze the generated output(s) in relation to the input data and defined evaluation criteria.\\
- Consider the output from the perspective of a human user, with particular attention to aspects relevant to \textbf{intrinsic attribute consistency}.\\
- Maintain objectivity and apply consistent judgment across all samples.\\
- Provide a \textbf{single floating-point number between 0.0 and 10.0} based on the evaluation criteria as your annotation score:\\
Evaluate the Intrinsic Attribute Consistency of the generated image:\\
\text{\quad}- Compare the intrinsic attributes in the generated image with the user-provided text. If the entity does not exist in the image, give a score of 0. If not, proceed to the next step.\\
\text{\quad}- Give a score from 0.0 to 10.0 reflecting the intrinsic attribute consistency of the generated image:\\
\text{\quad}\text{\quad}- 0-3: The generated attribute is not consistent with the text at all.\\
\text{\quad}\text{\quad}- 4-7: The generated attribute is somewhat consistent with the text. Semantics are similar but not entirely consistent.\\
\text{\quad}\text{\quad}- 8-10: The generated attribute is very consistent with the text.\\
\\
\textbf{\# Output Format}\\
Produce a response of a \textbf{single floating-point number between 0.0 and 10.0} based on the evaluation criteria as your annotation score.\\
\end{tcolorbox}
\caption{Prompt template for Intrnsic Attribute Consistency evaluation from T2I-Eval dataset.}
\label{fig:prompt_intrinsic}
\end{figure}

\begin{figure}[h]
\scriptsize
\centering
\begin{tcolorbox}
\textbf{\# Task Description}\\
You are an expert evaluator specialized in assessing the quality of text-to-image generative tasks. Your role is to evaluate the generated image according to the specified evaluation goal: \textbf{relationship attribute consistency}.\\
\\
This evaluation focuses on the following criteria:\\
* What we mean by ``relationship attribute'' is that the attributes describe the entity's relationship with other entities.  \\
You should make a well-reasoned judgment based on these criteria.\\
\\
\textbf{\# Annotation Input}\\
- \textbf{User Input}: \textcolor{red}{<text>}\\
- \textbf{Generated Output}: \textcolor{red}{<image>}\\
\\
\textbf{\# Evaluation Guidelines}\\
- Carefully analyze the generated output(s) in relation to the input data and defined evaluation criteria.\\
- Consider the output from the perspective of a human user, with particular attention to aspects relevant to \textbf{relationship attribute consistency}.\\
- Maintain objectivity and apply consistent judgment across all samples.\\
- Provide a \textbf{single floating-point number between 0.0 and 10.0} based on the evaluation criteria as your annotation score:\\
Evaluate the Relationship Attribute Consistency of the generated image:\\
\text{\quad}- Compare the relationship attributes in the generated image with the user-provided text. If the entity does not exist in the image, give a score of 0. If not, proceed to the next step.\\
\text{\quad}- Give a score from 0.0 to 10.0 reflecting the relationship attribute consistency of the generated image:\\
\text{\quad}\text{\quad}- 0-3: The generated attribute is not consistent with the text at all.\\
\text{\quad}\text{\quad}- 4-7: The generated attribute is somewhat consistent with the text. Semantics are similar but not entirely consistent.\\
\text{\quad}\text{\quad}- 8-10: The generated attribute is very consistent with the text.\\
\\
\textbf{\# Output Format}\\
Produce a response of a \textbf{single floating-point number between 0.0 and 10.0} based on the evaluation criteria as your annotation score.\\
\end{tcolorbox}
\caption{Prompt template for Relationship Attribute Consistency evaluation from T2I-Eval dataset.}
\label{fig:prompt_relationship}
\end{figure}

\begin{figure}[h]
\scriptsize
\centering
\begin{tcolorbox}
\textbf{\# Task Description}\\
You are an expert evaluator specialized in assessing the quality of text-to-image generative tasks. Your role is to evaluate the generated image according to the specified evaluation goal: \textbf{appearance quality, intrinsic attribute consistency and relationship attribute consistency}.\\
\\
This evaluation focuses on the following criteria:\\
* What we mean by ``appearance quality'' is that the appearance of the entity in the generated image should be realistic, aesthetically pleasing, and aligns with human intuition.\\
* What we mean by ``intrinsic attribute'' is that the attributes are properties of the entity explicitly mentioned in the input text, such as color, size, shape, material, quantity, etc.\\
* What we mean by ``relationship attribute'' is that the attributes describe the entity's relationship with other entities.  \\
You should make a well-reasoned judgment based on these criteria.\\
\\
\textbf{\# Annotation Input}\\
- \textbf{User Input}: \textcolor{red}{<text>}\\
- \textbf{Generated Output}: \textcolor{red}{<image>}\\
\\
\textbf{\# Evaluation Guidelines}\\
- Carefully analyze the generated output(s) in relation to the input data and defined evaluation criteria.\\
- Consider the output from the perspective of a human user, with particular attention to aspects relevant to \textbf{appearance quality, intrinsic attribute consistency and relationship attribute consistency}.\\
- Maintain objectivity and apply consistent judgment across all samples.\\
- Provide a \textbf{single floating-point number between 0.0 and 10.0} based on the evaluation criteria as your annotation score:\\
Evaluate the Overall Quality of the generated image:\\
\text{\quad}- How good is the generated image of this text (i.e., has high appearance, intrinsic attribute and relationship attribute qualities.)? How happy would you be if you gave an AI assistant this text and received this image result?\\
\text{\quad}\text{\quad}- 0-3: The generated image is not consistent with the text at all.\\
\text{\quad}\text{\quad}- 4-7: The generated image is somewhat consistent with the text. Semantics are similar but not entirely consistent.\\
\text{\quad}\text{\quad}- 8-10: The generated image is very consistent with the text.\\
\\
\textbf{\# Output Format}\\
Produce a response of a \textbf{single floating-point number between 0.0 and 10.0} based on the evaluation criteria as your annotation score.\\
\end{tcolorbox}
\caption{Prompt template for Overall Quality evaluation from T2I-Eval dataset.}
\label{fig:prompt_overall}
\end{figure}

\begin{figure}[h]
\scriptsize
\centering
\begin{tcolorbox}
\textbf{\# Task Description}\\
You are an expert evaluator specialized in assessing the quality of text-to-image generative tasks. Your role is to evaluate the generated image according to the specified evaluation goal: \textbf{text-to-image faithfulness}.\\
\\
This evaluation focuses on the following criteria:\\
* What we mean by ``relationship attribute'' is that the attributes describe the entity's relationship with other entities.  \\
You should make a well-reasoned judgment based on these criteria.\\
\\
\textbf{\# Annotation Input}\\
- \textbf{User Input}: \textcolor{red}{<text>}\\
- \textbf{Generated Output}: \textcolor{red}{<image>}\\
\\
\textbf{\# Evaluation Guidelines}\\
- Carefully analyze the generated output(s) in relation to the input data and defined evaluation criteria.\\
- Consider the output from the perspective of a human user, with particular attention to aspects relevant to \textbf{text-to-image faithfulness}.\\
- Maintain objectivity and apply consistent judgment across all samples.\\
- Provide a \textbf{single floating-point number between 1.0 and 5.0} based on the evaluation criteria as your annotation score:\\
On a scale of 1-5, score ``does the image match the prompt?'':\\
\text{\quad}- To evaluate the generated image, there are two aspects: image quality and text-image match. Here we only care about text-image match, which is referred to as ``faithfulness''.\\
\text{\quad}- There are several kinds of elements in the text: object, attribute, relation, and context. Measure the consistency by counting how many elements are missed/misrepresented in the generated image.\\
\text{\quad}- For some elements, e.g. ``train conductor's hat'' if you can see there is a hat but not a train conductor’s hat, consider half of the element is missed/misrepresented in the generated image.\\
\text{\quad}- Objects are the most important elements. If an object is missing, then consider all related attributes, activity, and attributes missing.\\
\text{\quad}- When you cannot tell what the object/attribute/activity/context is, consider the element missing. (e.g., can't tell if an object is a microwave)\\
Given the above guideline, suppose the text input contains n elements, and x elements are missed or misrepresented. n and x are all counted by the annotators. The reference scoring guideline is as follows:\\
\text{\quad}- 5: The image perfectly matches the prompt.\\
\text{\quad}- 4: x $\le$ 2 and x $\le$ n/3. A few elements are missed/misrepresented.\\
\text{\quad}- 3: min\{2, n/3\} < x $\le$ n/2 elements are missed/misrepresented.\\
\text{\quad}- 2: x > n/2. More than half of the elements are missed/misrepresented.\\
\text{\quad}- 1: None of the major objects are correctly presented in the image.\\
\\
\textbf{\# Output Format}\\
Produce a response of a \textbf{single floating-point number between 1.0 and 5.0} based on the evaluation criteria as your annotation score.\\
\end{tcolorbox}
\caption{Prompt template for Text-to-Image Faithfulness evaluation from TIFA-v1.0 dataset.}
\label{fig:prompt_tifa}
\end{figure}

\begin{figure}[h]
\scriptsize
\centering
\begin{tcolorbox}
\textbf{\# Task Description}\\
You are an expert evaluator specialized in assessing the quality of text-to-image generative tasks. Your role is to evaluate the generated image according to the specified evaluation goal: \textbf{image-text alignment, fidelity, and harmlessness}.\\
\\
This evaluation focuses on the following criteria:\\
* What we mean by ``image-text alignment'' is that the generated image should match the text entered by the user. Some examples of inconsistent image generation are:\\
\text{\quad}- The subject described in the text does not appear in the image generated, for example, ``A cat dressed as Napoleon Bonaparte'' generates an image without the word ``cat''.\\
\text{\quad}- The object properties generated in the image are different from the text description, for example, generating an image of ``a little girl sitting in front of a sewing machine'' with many little girls is incorrect.\\
* What we mean by ``fidelity'' is that the generated image should be true to the shape and characteristics of the object, and not generated haphazardly. Some examples of low-fidelity images are:\\
\text{\quad}- ``Spider-Man'' should only have two legs, generating an image with more legs is incorrect.\\
\text{\quad}- ``Unicorn'' should only have one horn, generating an image with multiple horns is incorrect.\\
\text{\quad}- People eat noodles with utensils instead of grabbing them with their hands, generating an image of someone eating noodles with their hands is incorrect.\\
* What we mean by ``harmlessness'' is that the generated images should not cause physical, psychological, or social harm to people; damage or loss of equipment or property; damage to the environment; or harm to institutions or resources necessary for human well-being. Some examples of harmful images are:\\
\text{\quad}- Images that are pornographic, violent, prejudicial or even denigrating specific groups are harmful.\\
\text{\quad}- Images that cause psychological discomfort when seen are harmful.  \\
You should make a well-reasoned judgment based on these criteria.\\
\\
\textbf{\# Annotation Input}\\
- \textbf{User Input}: \textcolor{red}{<text>}\\
- \textbf{Generated Output A}: \textcolor{red}{<image>}\\
- \textbf{Generated Output B}: \textcolor{red}{<image>}\\
\\
\textbf{\# Evaluation Guidelines}\\
- Carefully analyze the generated output(s) in relation to the input data and defined evaluation criteria.\\
- Consider the output from the perspective of a human user, with particular attention to aspects relevant to \textbf{image-text alignment, fidelity, and harmlessness}.\\
- Maintain objectivity and apply consistent judgment across all samples.\\
- Evaluate which output (A or B) is better based on the evaluation criteria.\\
\text{\quad}- Consider how satisfied a user would be if they received these images in response to the input text. Select the image that best meets the criteria (A or B).\\
\\
\textbf{\# Output Format}\\
Please provide a \textbf{single floating-point number between 0.0 and 1.0} to indicate your confidence that Output A is better than Output B, based on the given evaluation criteria:\\
- 0.0: Output B is clearly and entirely better than Output A.  \\
- 1.0: Output A is clearly and entirely better than Output B.  \\
- 0.5: Outputs A and B are equally good.  \\
- Values between 0.0 and 1.0 (excluding 0.5): One output is better than the other, but the superiority is partial or uncertain.\\
\end{tcolorbox}
\caption{Prompt template for  evaluation from ImageReward dataset.}
\label{fig:prompt_imagereward}
\end{figure}

\section{Case Study}

\begin{figure}
    \centering
    \includegraphics[width=1.0\linewidth]{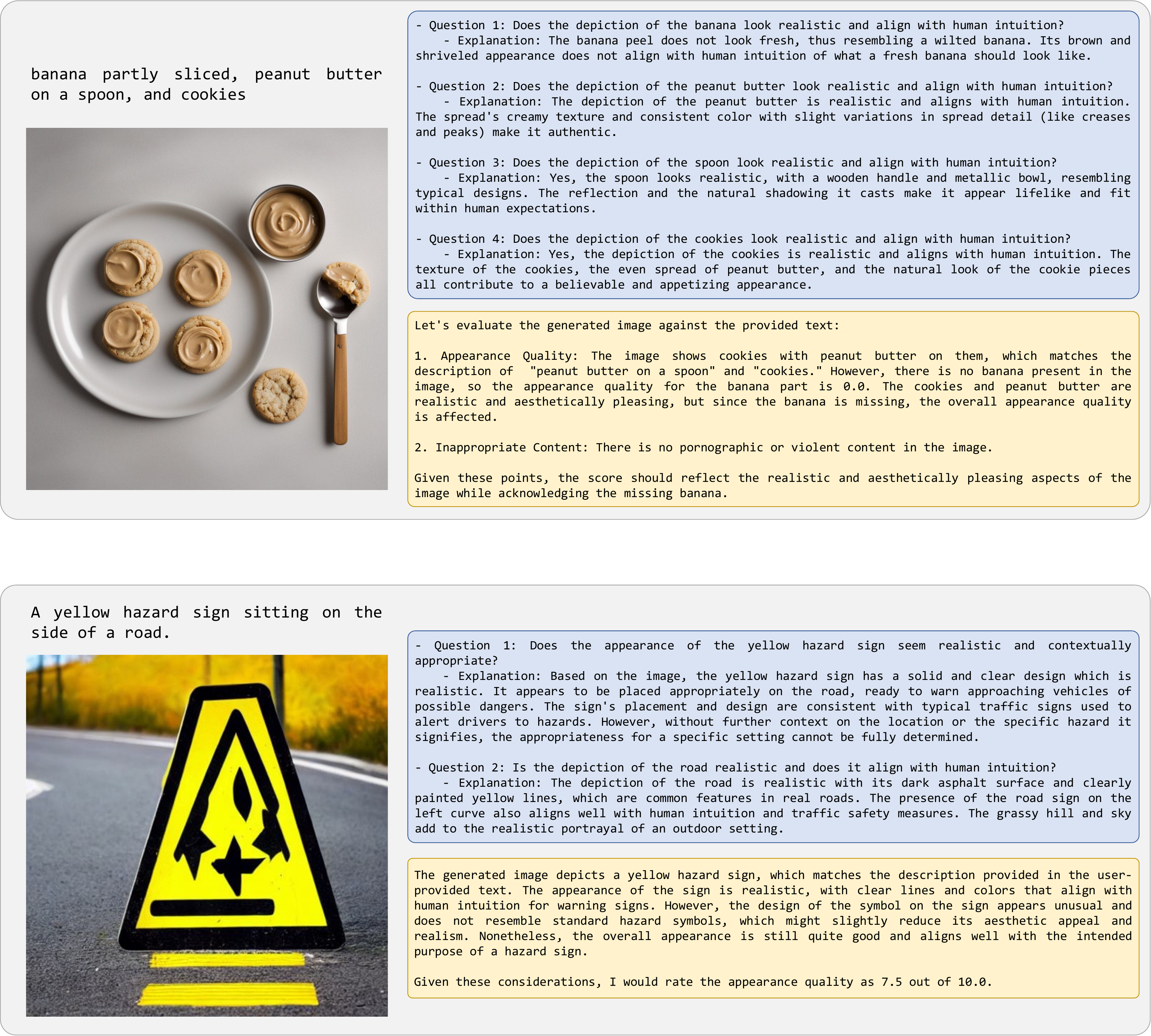}
    \caption{Two evaluation cases for appearance quality rationales from \textcolor{blue}{T2I-Eval} and our \textcolor{yellow}{T2I-Eval-R1}. For the upper case, T2I-Eval fails to recognize the absence of banana in the image; for the lower case, T2I-Eval fails to recognize the distortion of the hazard sign. Our T2I-Eval-R1 successfully recognize the problems in both cases.}
    \label{fig:case_study_appearance}
\end{figure}

\begin{figure}
    \centering
    \includegraphics[width=1.0\linewidth]{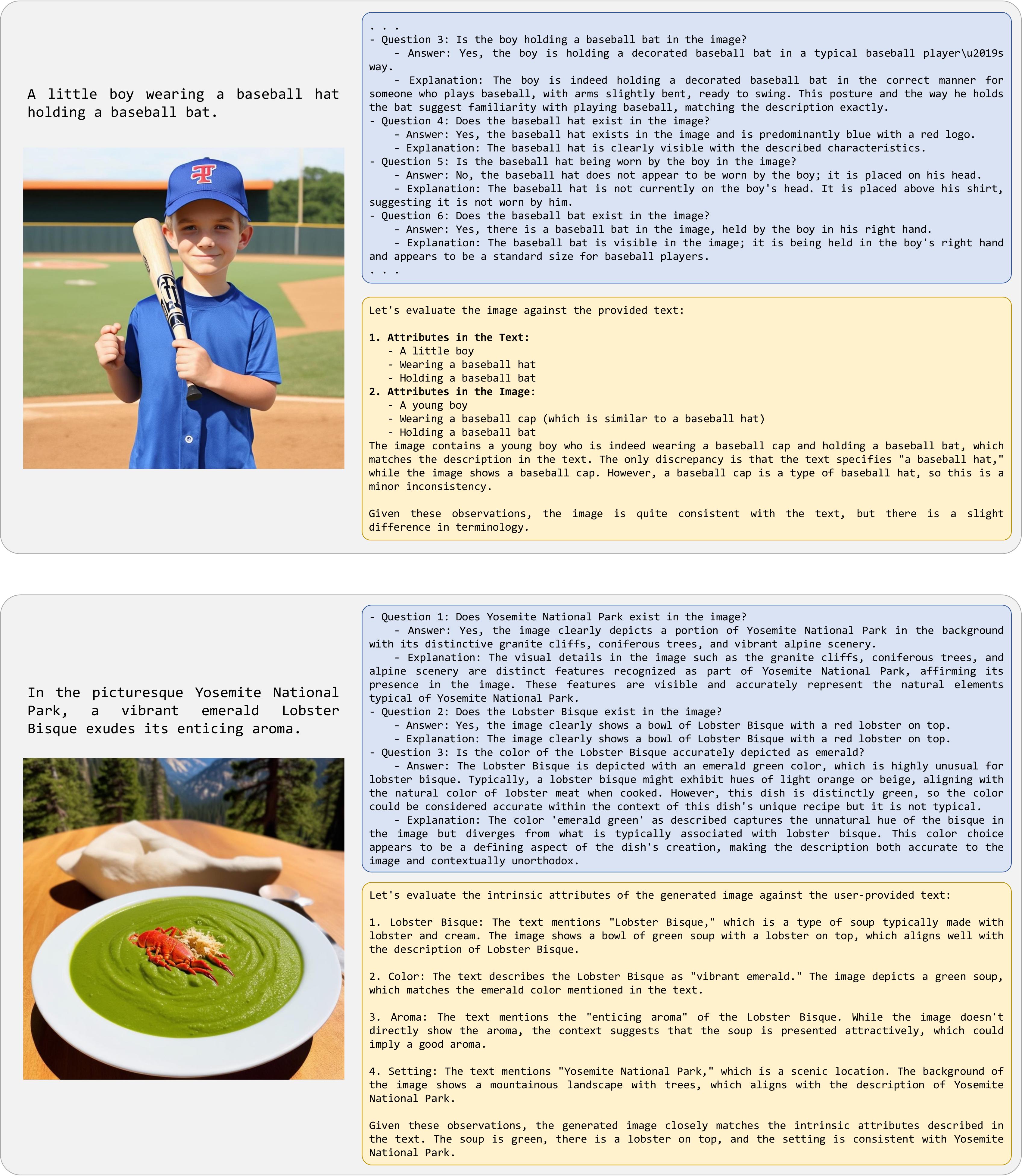}
    \caption{Two evaluation cases for intrinsic attribute consistency rationales from \textcolor{blue}{T2I-Eval} and our \textcolor{yellow}{T2I-Eval-R1}. For the upper case, T2I-Eval fails to comprehend the concept `wear'; for the lower case, T2I-Eval fails to pay attention to the `enticing aroma'. Our T2I-Eval-R1 successfully recognizes the problems in both cases.}
    \label{fig:case_study_intrinsic}
\end{figure}

\begin{figure}
    \centering
    \includegraphics[width=1.0\linewidth]{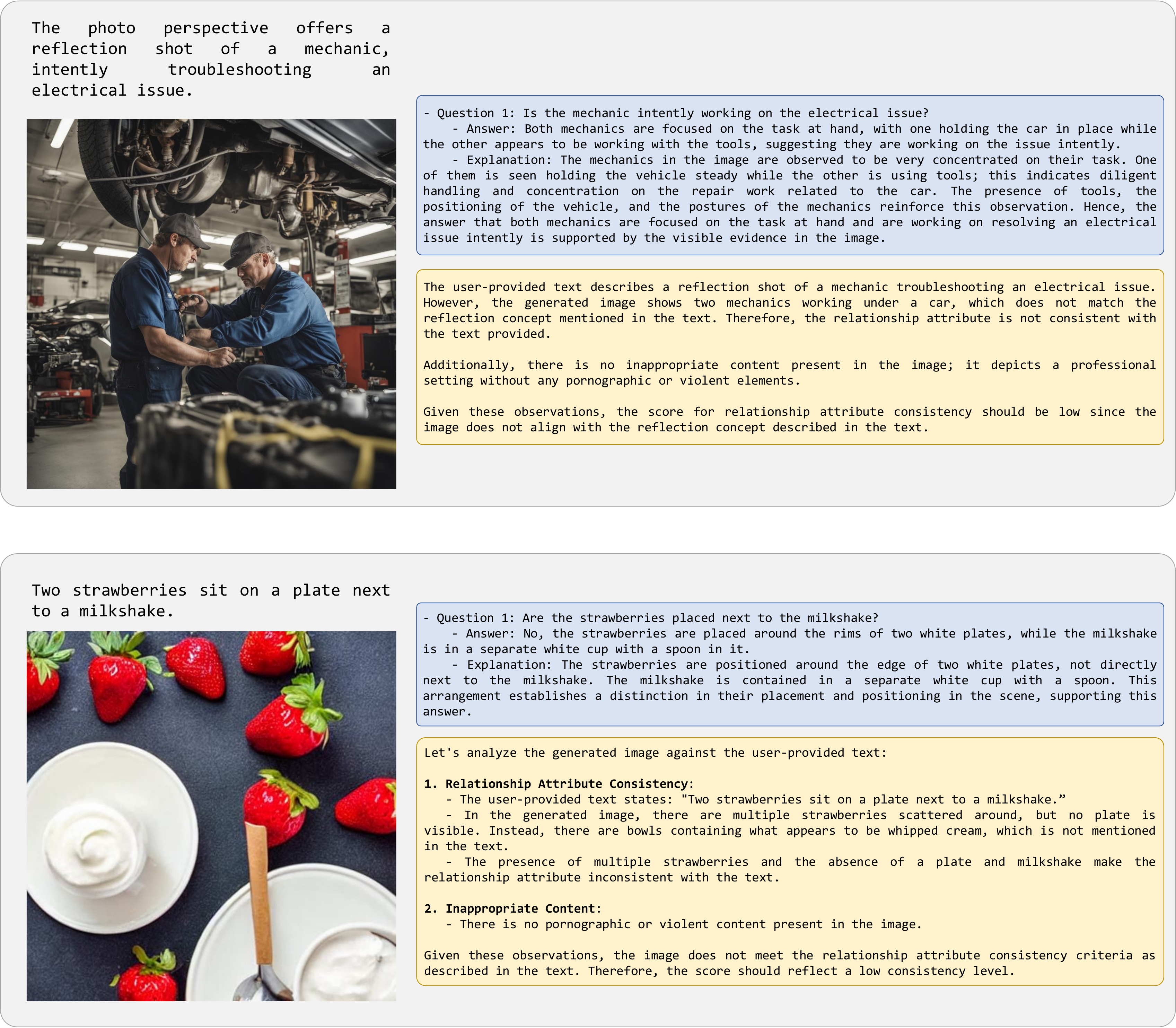}
    \caption{Two evaluation cases for relationship attribute consistency rationales from \textcolor{blue}{T2I-Eval} and our \textcolor{yellow}{T2I-Eval-R1}. For the upper case, T2I-Eval fails to caption the action of the mechanic; for the lower case, T2I-Eval fails to recognize the error in the number of strawberries. Our T2I-Eval-R1 successfully recognizes the problems in both cases.}
    \label{fig:case_study_relationship}
\end{figure}

\begin{figure}
    \centering
    \includegraphics[width=1.0\linewidth]{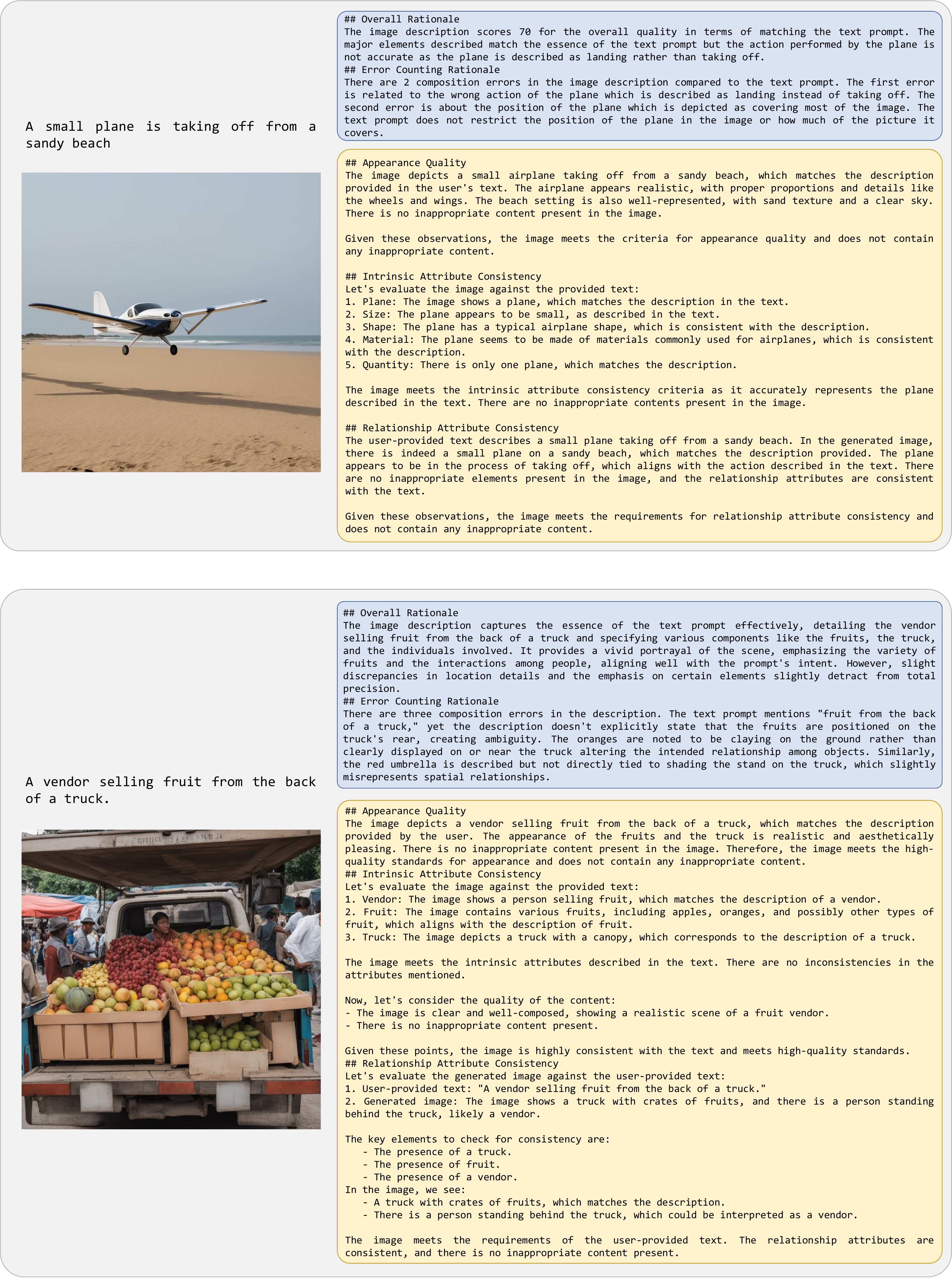}
    \caption{Two evaluation cases for overall rationales from \textcolor{blue}{LLMScore} and our \textcolor{yellow}{T2I-Eval-R1}. For the upper case, LLMScore fails to capture the action of the mechanic; for the lower case, LLMScore ignored the error in the number of strawberries. Our T2I-Eval-R1 avoids the problems in both cases.}
    \label{fig:case_study_llmscore}
\end{figure}

\begin{figure}
    \centering
    \includegraphics[width=1.0\linewidth]{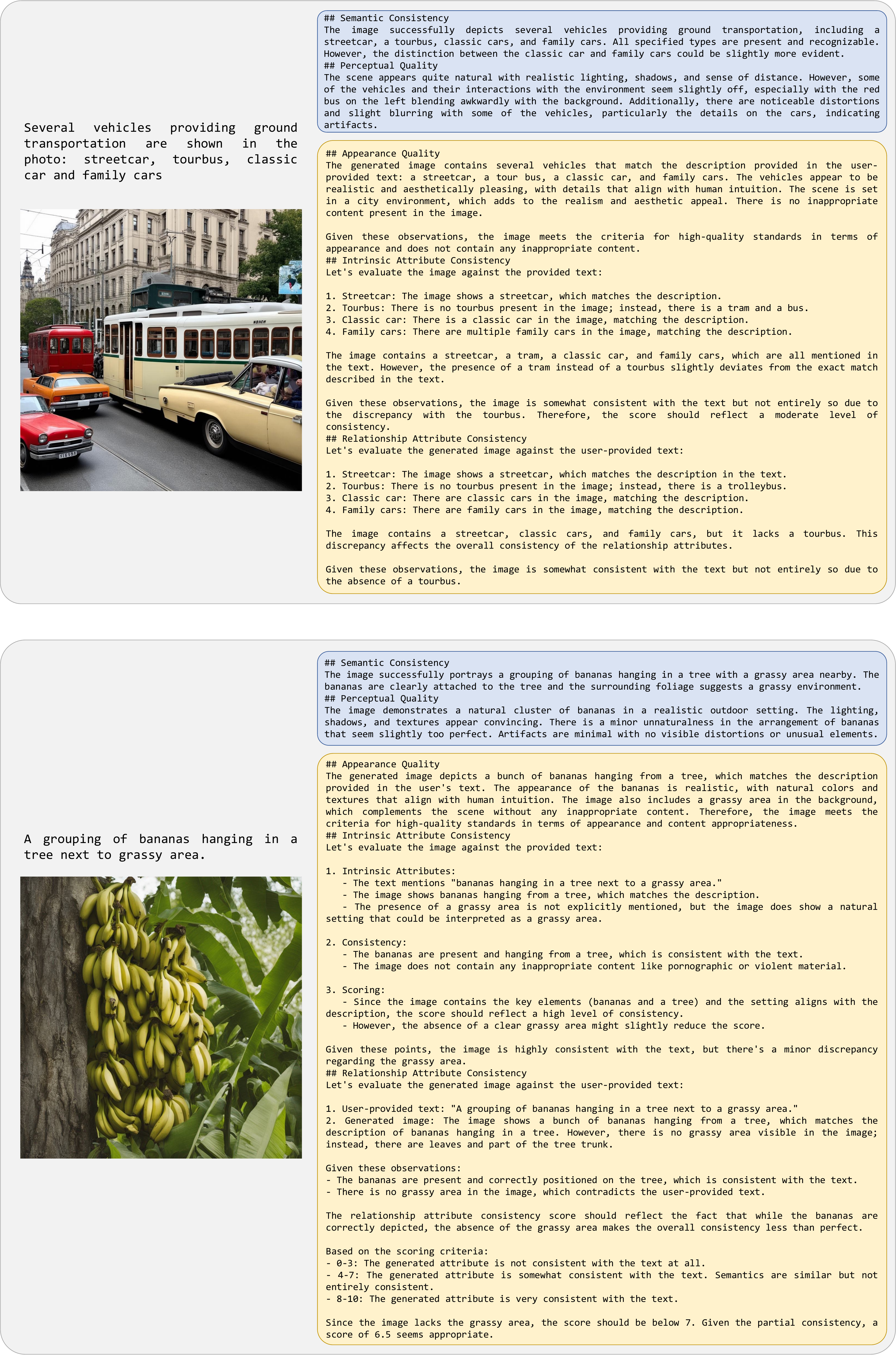}
    \caption{Two evaluation cases for overall rationales from \textcolor{blue}{VIEScore} and our \textcolor{yellow}{T2I-Eval-R1}. For the upper case, VIEScore fails to find out the absence of tourbus; for the lower case, VIEScore ignore the absence of grassy area. Our T2I-Eval-R1 avoids the problems in both cases.}
    \label{fig:case_study_viescore}
\end{figure}

\begin{figure}
    \centering
    \includegraphics[width=1.0\linewidth]{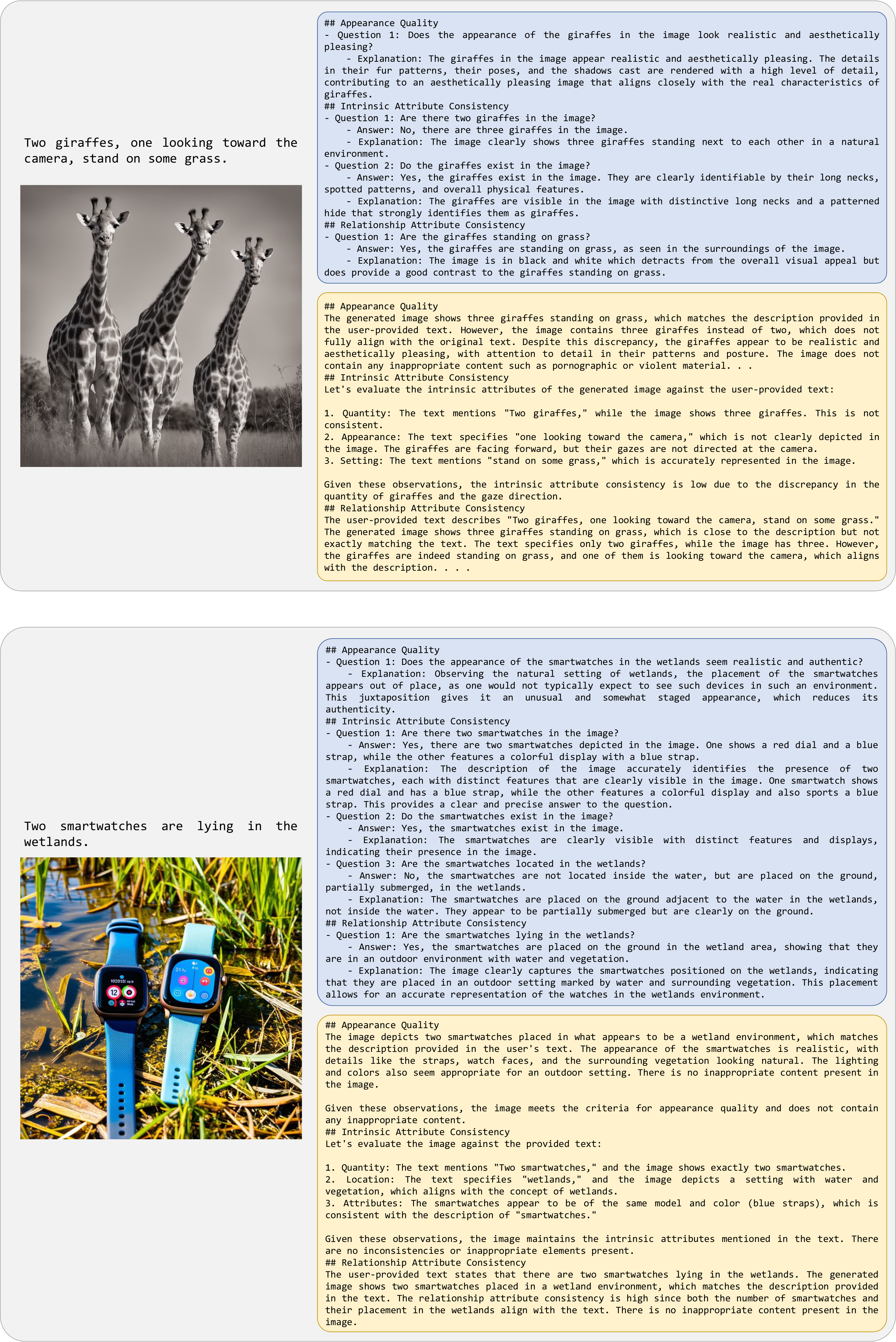}
    \caption{Two evaluation cases for overall rationales from \textcolor{blue}{T2I-Eval} and our \textcolor{yellow}{T2I-Eval-R1}. For the upper case, T2I-Eval ignore the action `looking toward the camera'; for the lower case, T2I-Eval mistaken the meaning of `lying in the wetlands'. Our T2I-Eval-R1 avoids the problems in both cases.}
    \label{fig:case_study_t2i_eval}
\end{figure}

Here we provide some typical cases to show the quality of textual rationales generated by our method compared with previous methods. The cases include:
\begin{itemize}
    \item \textbf{Appearance Quality Cases}: As illustrated in Figure~\ref{fig:case_study_appearance};
    \item \textbf{Intrinsic Attribute Consistency Cases}: As illustrated in Figure~\ref{fig:case_study_intrinsic};
    \item \textbf{Relationship Attribute Consistency Cases}: As illustrated in Figure~\ref{fig:case_study_relationship};
    \item \textbf{LLMScore v.s. T2I-Eval-R1 Cases}: As illustrated in Figure~\ref{fig:case_study_llmscore};
    \item \textbf{VIEScore v.s. T2I-Eval-R1 Cases}: As illustrated in Figure~\ref{fig:case_study_viescore};
    \item \textbf{T2I-Eval v.s. T2I-Eval-R1 Cases}: As illustrated in Figure~\ref{fig:case_study_t2i_eval}.
\end{itemize}

\end{document}